\documentclass[sigconf]{acmart}
\AtBeginDocument{%
  \providecommand\BibTeX{{%
    \normalfont B\kern-0.5em{\scshape i\kern-0.25em b}\kern-0.8em\TeX}}}

\usepackage{microtype}
\usepackage{csquotes}

\renewcommand{\baselinestretch}{0.959}
\DeclareMathOperator*{\argmax}{arg\,max}
\DeclareMathOperator*{\argmin}{arg\,min}

\usepackage{tabularx}
\usepackage{lscape}
\usepackage{balance}
\copyrightyear{2024}
\acmYear{2024}
\setcopyright{rightsretained}
\acmConference[HRI '24]{Proceedings of the 2024 ACM/IEEE International Conference on Human-Robot Interaction}{March 11--14, 2024}{Boulder, CO, USA}
\acmBooktitle{Proceedings of the 2024 ACM/IEEE International Conference on Human-Robot Interaction (HRI '24), March 11--14, 2024, Boulder, CO, USA}
\acmDOI{10.1145/3610977.3634964}
\acmISBN{979-8-4007-0322-5/24/03}

\begin{document}
%\title{Independence in the Home: A Wearable Interface for a Person with Quadriplegia to Teleoperate a Mobile Manipulator}

\title{Independence in the Home: A Wearable Interface for a Person with Quadriplegia to Teleoperate a Mobile Manipulator}
\author{Akhil Padmanabha}
\affiliation{
  %\institution{Robotics Institute}
  \institution{Carnegie Mellon University}
  \city{Pittsburgh}
  \state{PA}
  \country{USA}
}
\email{akhilpad@andrew.cmu.edu}

\author{Janavi Gupta}
\affiliation{
  %\institution{School of Computer Science}
  \institution{Carnegie Mellon University}
  \city{Pittsburgh}
  \state{PA}
  \country{USA}
}
\email{janavig@andrew.cmu.edu}

\author{Chen Chen}
\affiliation{
  %\institution{Department of Automation}
  \institution{Tsinghua University}
  \city{Beijing}
  \country{China}
}
\email{chen-che20@mails.tsinghua.edu.cn}

\author{Jehan Yang}
\affiliation{
  %\institution{Department of Biomedical Engineering}
  \institution{Carnegie Mellon University}
  \city{Pittsburgh}
  \state{PA}
  \country{USA}
}
\email{jehan@cmu.edu}

\author{Vy Nguyen}
\affiliation{
  \institution{Hello Robot Inc.}
  \city{Martinez}
  \state{CA}
  \country{USA}
}
\email{v@hello-robot.com}

\author{Douglas J. Weber}
\affiliation{
  %\institution{Department of Mechanical Engineering}
  \institution{Carnegie Mellon University}
  \city{Pittsburgh}
  \state{PA}
  \country{USA}
}
\email{dweber2@andrew.cmu.edu}

\author{Carmel Majidi}
\affiliation{
  %\institution{Department of Mechanical Engineering}
  \institution{Carnegie Mellon University}
  \city{Pittsburgh}
  \state{PA}
  \country{USA}
}
\email{cmajidi@andrew.cmu.edu }

\author{Zackory Erickson}
\affiliation{
  %\institution{Robotics Institute}
  \institution{Carnegie Mellon University}
  \city{Pittsburgh}
  \state{PA}
  \country{USA}
}
\email{zackory@cmu.edu}

\newcommand\blfootnote[1]{
    \begingroup
    \renewcommand\thefootnote{}\footnote{#1}
    \addtocounter{footnote}{-1}
    \endgroup
}

\renewcommand{\shortauthors}{Akhil Padmanabha et al.}




\begin{abstract}
 \vspace{-0.1cm}
  Teleoperation of mobile manipulators within a home environment can significantly enhance the independence of individuals with severe motor impairments, allowing them to regain the ability to perform self-care and household tasks. There is a critical need for novel teleoperation interfaces to offer effective alternatives for individuals with impairments who may encounter challenges in using existing interfaces due to physical limitations. In this work, we iterate on one such interface, HAT (Head-Worn Assistive Teleoperation), an inertial-based wearable integrated into any head-worn garment. We evaluate HAT through a 7-day in-home study with Henry Evans, a non-speaking individual with quadriplegia who has participated extensively in assistive robotics studies. We additionally evaluate HAT with a proposed shared control method for mobile manipulators termed Driver Assistance and demonstrate how the interface generalizes to other physical devices and contexts. Our results show that HAT is a strong teleoperation interface across key metrics including efficiency, errors, learning curve, and workload. Code and videos are located on our project website\footnote{\url{https ://sites.google.com/view/hat2-teleop/}}.
  \vspace{-0.15cm}
 \blfootnote{This research is supported by the National Science Foundation Graduate Research Fellowship Program under Grant No. DGE1745016 and DGE2140739.}
\end{abstract}

\begin{CCSXML}
<ccs2012>       <concept_id>10010520.10010553.10010554.10010558</concept_id>
       <concept_desc>Computer systems organization~External interfaces for robotics</concept_desc>
       <concept_significance>500</concept_significance>
       </concept>
   <concept>
       <concept_id>10010520.10010553.10010554.10010555</concept_id>
       <concept_desc>Computer systems organization~Robotic components</concept_desc>
       <concept_significance>500</concept_significance>
       </concept>
 </ccs2012>
\end{CCSXML}

\ccsdesc[500]{Computer systems organization~External interfaces for robotics}
\ccsdesc[500]{Computer systems organization~Robotic components}

\keywords{assistive robotics, teleoperation, in-the-wild studies, shared control}

\begin{teaserfigure}
  \vspace{-0.35cm}
  \includegraphics[width=\textwidth]{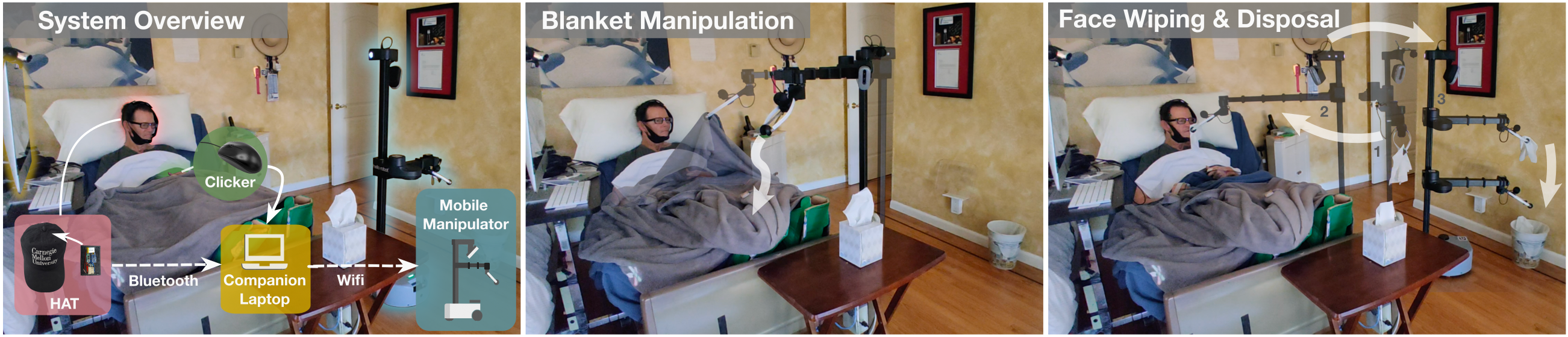}
  \vspace{-0.7cm}
  \caption{Henry, a non-speaking individual with quadriplegia, uses HAT, a head-worn assistive interface, to teleoperate a mobile manipulator to move a blanket below his knees, grasp a tissue to wipe his face, and throw the tissue in a trash can. The interface records and wirelessly sends Henry's head orientation angles to a companion laptop, which computes and transmits actuator velocities to the robot. A clicker is used by Henry to switch between modes, distinct operational states of the HAT system.}
  \Description{Henry, who is a non-speaking individual with quadriplegia, uses HAT, a head-worn assistive interface for mobile manipulators. The first image provides a system overview. The interface records and sends Henry's head orientation angles over Bluetooth to a companion laptop, which computes and transmits actuator velocities to the robot over WiFi. A clicker, plugged into the companion laptop, is used by Henry to switch between modes, distinct operational states of the HAT system. The second and third images shows snippets from a task conducted during the study. In the second image, Henry uses the interface to move a blanket below his knees. In the third image, Henry uses HAT to grasp a tissue to wipe his face and throw the tissue in a trash can.}
  \vspace{-0.05cm}
  \label{fig:teaser}
\end{teaserfigure}


\maketitle

%\email{email}
\vspace{-0.15cm}
\section{Introduction}
\vspace{-0.1cm}
Impairments in motor function such as the inability to use an upper extremity can severely limit a person's ability to engage in both activities of daily living (ADLs) such as bathing, eating, and dressing and instrumental activities of daily living (iADLs) such as home management (cleaning, meal preparation, etc). Additionally, motor impairments can significantly impact an individual’s physical capacity and independence to engage in leisure activities like playing cards, choosing TV channels, participating in social activities such as interacting with family and friends, and more. Many people who experience such motor limitations must depend on caregivers for assistance, leading to measureable impacts in quality of life and independence~\cite{qol1, qol2, qol3, qol4}. This problem is widespread with motor impairments affecting a significant portion of the United States population, with approximately 5 million individuals (1.7\%), dealing with varying degrees of paralysis due to neurodegenerative conditions, stroke, muscle degradation, or spinal cord injuries (SCI)~\cite{armour2016prevalence}.  While these individuals might experience motor impairments, many retain the ability to move their head and neck~\cite{world2013international}. For those with quadriplegia or limited hand dexterity, technologies that enable completion of both ADLs/iADLs and leisure tasks can be transformative~\cite{impact1, impact2, impact3, impact4, impact5}. 

Specifically, teleoperation of mobile manipulators in the home can empower individuals with motor impairments, enabling them to complete various self-care and household tasks~\cite{tasks, tasksaroundhead,king2012dusty, robotsforhumanity, yang2023high}. While developing teleoperation interfaces, it is crucial to consider metrics such as physical/mental workload, ease of use, and learning curve. Most interfaces come with physical prerequisites, essentially determining the subset of individuals with impairments who can effectively utilize them. For instance, hand-operated joysticks offer precise and continuous control but necessitate functional hand motor skills. Conversely, web interfaces depend on devices with screens, requiring users to manipulate a cursor and interact with buttons, often demanding fine cursor control through an assistive device such as head tracking~\cite{headtracking, headtracking2, FaceMouse} or eye tracking~\cite{jacob1995eye, kaufman1993eye}. Developing and evaluating novel assistive interfaces for mobile manipulators can present preferable alternatives for individuals with impairments who may struggle to use conventional options.

% Head gestures are a generalized behavior used by individuals to communicate more effectively during conversation~\cite{mehrabian2017communication} and could be extended to express human intent to a mobile manipulator. 

%Lastly, brain computer interfaces (BCI) show promise for efficient control of physical devices for individuals with motor impairments. Researchers have demonstrated control of a robotic arm for tetraplegics using embedded neural interfaces~\cite{BrainGate1}. Even though potentially applicable to a large population, these BCI techniques are often invasive and need to be extensively calibrated with each new user. As an alternative to BCI, EEG and EMG signals from skin-mounted electrodes are non-invasive and have been used to teleoperate simple robot motions based on residual myoelectric signals~\cite{emg1, emg2}. In contrast, a head-worn noninvasive assistive interface may be generally applicable to a wide population of individuals who have lost significant motor function below the neck, as is the case for many individuals with cervical SCI or tetraplegia due to neurodegenerative disease, stroke, or injury.

In this work, we build on our existing, inertial measurement unit (IMU) based interface, HAT (Head-Worn Assistive Teleoperation), which can be embedded in any head-worn clothing article such as a baseball cap~\cite{padmanabha2023hat}. HAT serves as a direct teleoperation interface which maps head orientation angles to actuator velocities for a high degree of freedom mobile manipulator, Hello Robot's Stretch~\cite{stretch}. HAT can be used by individuals with head motion, which can often be retained after cervical spinal cord injury or quadriplegia~\cite{world2013international}. Research has suggested that head-worn interfaces for mobile manipulators may offer individuals with impairments a strong alternative compared to traditional web-based interfaces~\cite{padmanabha2023hat}. 

We note several limitations in head-worn control interfaces that we aim to address in this research. First, prior evaluations with HAT were conducted through 2-hour long studies primarily with non-disabled individuals in confined lab environments~\cite{padmanabha2023hat}. Thus, to evaluate longer term usage of the interface in a real-world setting, we conducted and present results from a 7 day in-home case study with Henry Evans, an individual with quadriplegia. Case studies (N=1) are a proven research methodology in the fields of medicine, human computer interaction, and human robot interaction, that can provide in-depth analysis in a particular process or technology~\cite{robotsforhumanity, foloppe2018potential, neuper2003clinical, yozbatiran2012robotic, metzger2023high}. Second, despite participants' impression of a head-worn interface as efficient~\cite{padmanabha2023hat}, task times were still much higher than if a caregiver or non-disabled individual completed the task. To reduce these long task times, overshooting errors, and effort for grasp alignment especially with reduced depth perception, we introduce Driver Assistance for mobile manipulators, inspired by driver assistance features in motorized vehicles. Driver Assistance, in our case, is a shared control method for precision grasping of household objects. Lastly, the need for customization in a wearable interface is paramount as impairments vary across a spectrum encompassing a wide range of conditions and physical abilities. Thus, we introduce multiple wearable designs, capture an additional degree of head motion for robot control, and integrate a clicker for mode switching. We further generalize these results by connecting HAT to a GUI that allows for robust robot control out of line-of-sight and for control of other physical devices in the home.

The contributions of this work are as follows: 
\vspace{-0.05cm}
\begin{itemize}
    \item We evaluate an inertial-based assistive interface, HAT, with a mobile manipulator in a week-long case study with an individual with quadriplegia in their home environment and assess a variety of metrics. 
    \item We introduce Driver Assistance, a form of shared control for mobile manipulators, and show that, for our participant, it reduces task times and effort while still preserving the perception of control.
    \item We demonstrate the customizability and generalizability of the interface and show how it can adapt to individuals with severe impairments, to different physical devices in the environment, and to different contexts.
\end{itemize}

\vspace{-0.1cm}
\section{Related Work}
\subsection{Teleoperation Interfaces}
Many forms of assistive interfaces, including hand/mouth operated joysticks, have been evaluated with mobile robots and wheelchairs~\cite{fehr2000adequacy, pazzaglia2016embodiment}. Web-based interfaces~\cite{robotsforhumanity, surrogates, TapoMaya, ranganeni2023evaluating}, the prevailing norm for teleoperation of robots, enable direct control of actuators using screen-based buttons and camera feeds. These interfaces can be used with a standard mouse but are also compatible with alternate assistive devices such as head tracking~\cite{headtracking, headtracking2, FaceMouse} and eye tracking~\cite{jacob1995eye, kaufman1993eye}. Over the last decade, numerous web-based teleoperation interfaces have been developed and evaluated with individuals with and without motor impairments~\cite{tasks,tasksaroundhead,robotsforhumanity, surrogates,TapoMaya}.

%Researchers have evaluated several iterations of these teleoperation platforms with both healthy participants and one participant with tetraplegia using a PR2 robot~\cite{tasks,tasksaroundhead,robotsforhumanity}. This interface enabled control of the robot for various tasks including grasping in cluttered environments, handing out candy, opening of a drawer and extracting an object, etc. Efforts have been made to make these interfaces more intuitive, by overlaying movement buttons directly over camera feeds and providing additional views of the environment, and to evaluate them with more individuals with motor impairments~\cite{surrogates,TapoMaya}. 

\vspace{-0.1cm}
\subsection{IMU-based Interfaces}
IMU-based interfaces have previously been explored for control of mobile robots~\cite{mobilerobotIMU, IMUglove}. While highly efficient, these interfaces cannot be used by individuals without intact limb movement. Head-worn IMU-based interfaces have been used in conjunction with electric wheelchairs~\cite{wheelchairIMU, wheelchair2, wheelchair3}, but most have not been evaluated with mobile manipulators that have many additional degrees of freedom and which require the user to teleoperate the mobile base in third person. HAT is an IMU-based, direct teleoperation interface which maps head orientation angles to actuator velocities for a high degree of freedom mobile manipulator~\cite{padmanabha2023hat}. In the past study, participants used head tilts along the pitch and roll axes to control robot velocities. They used speech to switch between modes, which dictated the subset of actuators they could control with their head motions. Researchers tested the interface with 16 non-disabled and 2 participants with motor impairments in 2-hour study sessions per participant. Participants generally found the interface efficient and error-recovery friendly. Ease of use and learning curve were rated positively, and most participants reported low workload measures.

\subsection{Shared Control}
\label{sc}
Shared control (SC) involves a robot adjusting its level of autonomy based on its comprehension of human intentions and the surrounding environment~\cite{selvaggio2021autonomy, brenna2018autonomy, hameed2023control}, resulting in a reduction in cognitive load for the user~\cite{merkt2017robust, losey2018review, gopinath2017human}. SC has been explored in diverse contexts, such as manipulators/humanoids~\cite{daniel2019shared,xi2019robotic,zhu2020human}, wheelchairs~\cite{erdogan2017effect,sinyukov2018}, and surgical robots~\cite{zhang2022human,xiong2017}. Research has also examined SC across varying levels of autonomy~\cite{hameed2023control}. Past works typically infer the user's intention via their input and provide assistance through control blending~\cite{losey2018review, jain2020probabilistic, selvaggio2021autonomy}. However, few SC methods have been tested in complex real-world environments, such as household settings. While developing SC methods, the user's feeling of control over the system is an important consideration as previous studies show that higher levels of autonomy can lead to a reduced feeling of control~\cite{javdani2018,bhattacharjee2020}. In comparison to past work, our proposed SC method, Driver Assistance, limits the shared control to specific robot joints to exemplify the human perception of control.

%SC methods have been explored for assistive teleoperation in robot arms and mobile manipulators~\cite{brenna2018autonomy,jain2020probabilistic,merkt2017robust}

\vspace{-0.1cm}
\section{Methodology}
We conducted an evaluation of the system with a single participant, Henry Evans (M, 62). Henry faces unique challenges as a non-speaking individual with quadriplegia due to brainstem stroke. He experiences some limitation in his neck's range of motion and he has slight flexion of his left thumb, allowing him to click a single button on a computer mouse. Henry has accumulated extensive robotics experience through his involvement in numerous studies with web-based interfaces~\cite{robotsforhumanity, kapusta2019system, surrogates, tasksaroundhead, tasks, Ackerman_2023, nguyenincreasing} and assistive robots. Our study, approved by Carnegie Mellon University's Institutional Review Board, spanned a period of seven consecutive days from 8/14/2023 to 8/20/2023 in Henry’s home environment. Recruitment was conducted over email and informed consent was obtained from Henry for use of name and image in publication. Before and throughout the study, we adhered to user-centered design principles, actively seeking input from Henry at various stages of system development. This process led to improvements to existing functionality and the creation of new features for HAT. 

%This study was approved by Carnegie Mellon University's Institutional Review Board. 

%Based on Henry's feedback, we first expanded on the interface's customizability by introducing two innovative wearable prototypes, incorporating a single-button clicker for mode switching, and allowing use of the yaw axis for robot control. We additionally introduced a cursor control mode, enabling Henry to navigate his computer using HAT. In response to limitations and input from reviewers of our past publication, we also developed a graphical user interface (GUI) with camera views to facilitate tasks that required operation outside the line-of-sight. Lastly, we introduced a shared autonomy feature called "driver assistance." 

\vspace{-0.1cm}
\subsection{Device Design}

As seen in Fig.~\ref{fig:teaser}, the system consists of HAT, a companion laptop, and Hello Robot's Stretch RE2 robot. HAT consists of a clothing article and a foam pad on which the electronics, specified in Appendix~\ref{device_electronics}, are mounted on. Velcro is glued to the pad and can be used to secure it to a head clothing article of the user's choosing. The IMU, attached such that it remains flat at the crown of the head, estimates and sends roll, pitch, and yaw absolute orientation angles at 20 Hz to the companion laptop over Bluetooth. 

We evaluated three different head-worn garments (baseball cap, cloth headband, and silicone chin strap) for this study. The chin strap, suggested by Henry and shown in Fig.~\ref{device_design}, is traditionally used to reduce snoring or minimize chin wrinkles. Henry proposed the chin strap for two reasons: 1. Its secure fit, preventing movement during spasms 2. Comfort due to its design without material at the back of the head. After testing, Henry chose to use the chin strap for the study with slight modifications to enlarge the ear holes. Throughout the week, we discovered two additional benefits: it prevented his glasses from slipping and effectively reduced drooling by keeping his mouth comfortably closed.

\subsection{Mode Switching}
\begin{figure}[t!]
      \centering
      \includegraphics[width = \columnwidth]{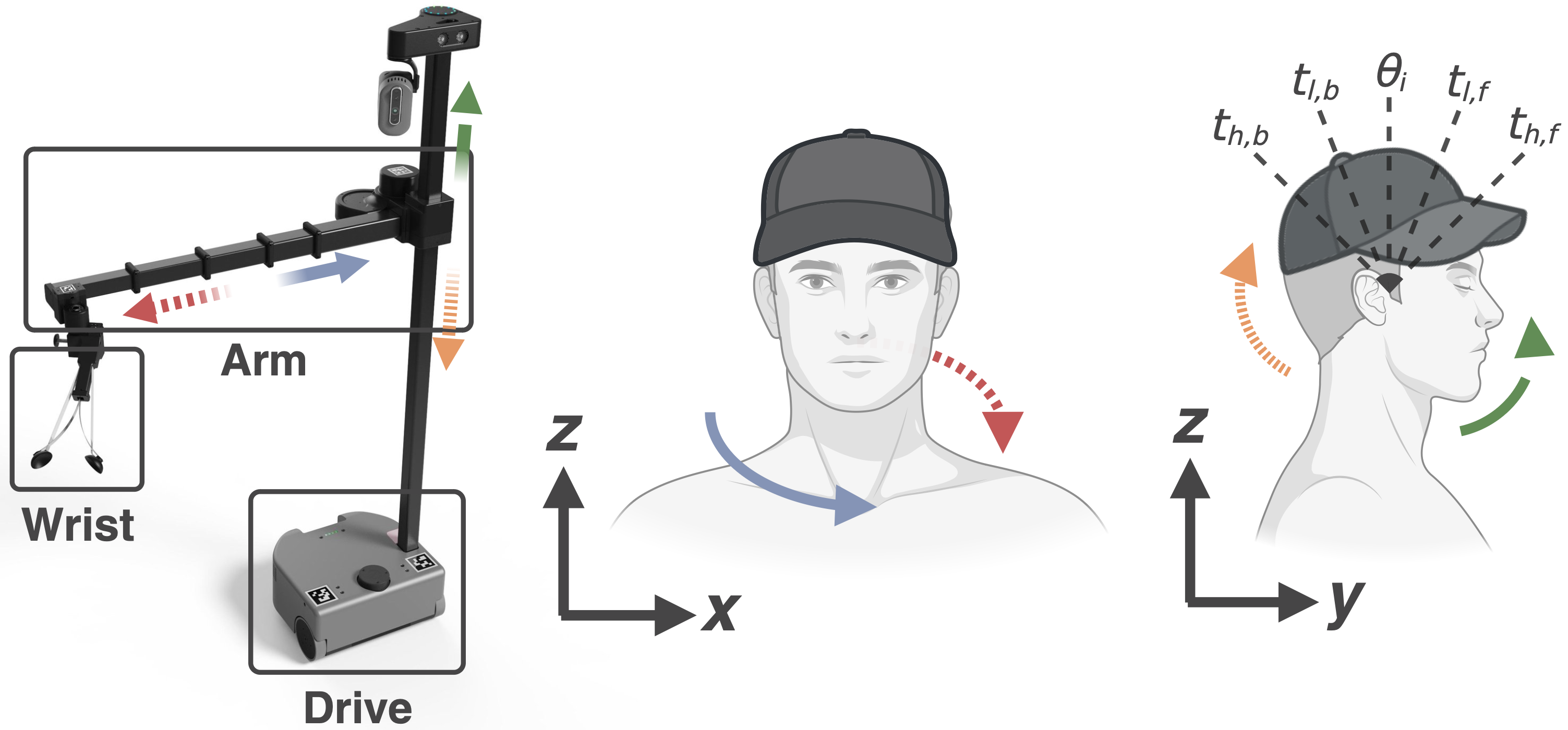}
      \vspace{-0.6cm}
      \caption{Three robot control modes: Drive, Arm, Wrist and mapping of head orientation to robot motion in arm mode.}
      \Description{On the left of the figure is Hello Robot's Stretch, the mobile manipulator utilized. The three modes, wrist, arm, and drive are labelled on the robot. On the right are two figures of a person wearing a hat, showing how head tilting maps to robot movement.}
      \vspace{-0.45cm}
      \label{robot_modes}
      \setcounter{equation}{1}
   \end{figure}

Mode switching involves transitioning between distinct operational states within the HAT system, each offering the user specific functionalities. For example, while in each of the robot control modes (drive, arm, and wrist) shown in Fig.~\ref{robot_modes}, Henry can command a subset of robot actuators with head motions. For this study, we integrated Henry's clicker, which is a standard computer mouse shown in Fig.~\ref{fig:teaser}, as he is non-speaking. The clicker is plugged directly into the companion laptop, which announces the mode change with an audio notification. We additionally added two modes, idle and cursor control, to expand the available functionality. In idle mode, the HAT is deactivated allowing the participant to move their head freely without causing robot motion. In cursor control mode, head motions can be used to control the cursor on the companion laptop. To switch between the modes, Henry can use 4 distinct clicking patterns: single click, double click, triple click, and holding down the clicker. We conducted two rounds of iteration on these clicking patterns during the study at the start of Day 3 and Day 6, incorporating Henry's feedback. The clicking patterns are described in more detail in Appendix~\ref{appendix_mode_switching}. The integration of an alternative mode-switching method and the ability to easily modify clicking patterns demonstrates the flexibility of the HAT system for customization.

\begin{figure*}[hbt!]
      \centering
      \includegraphics[width = \textwidth]{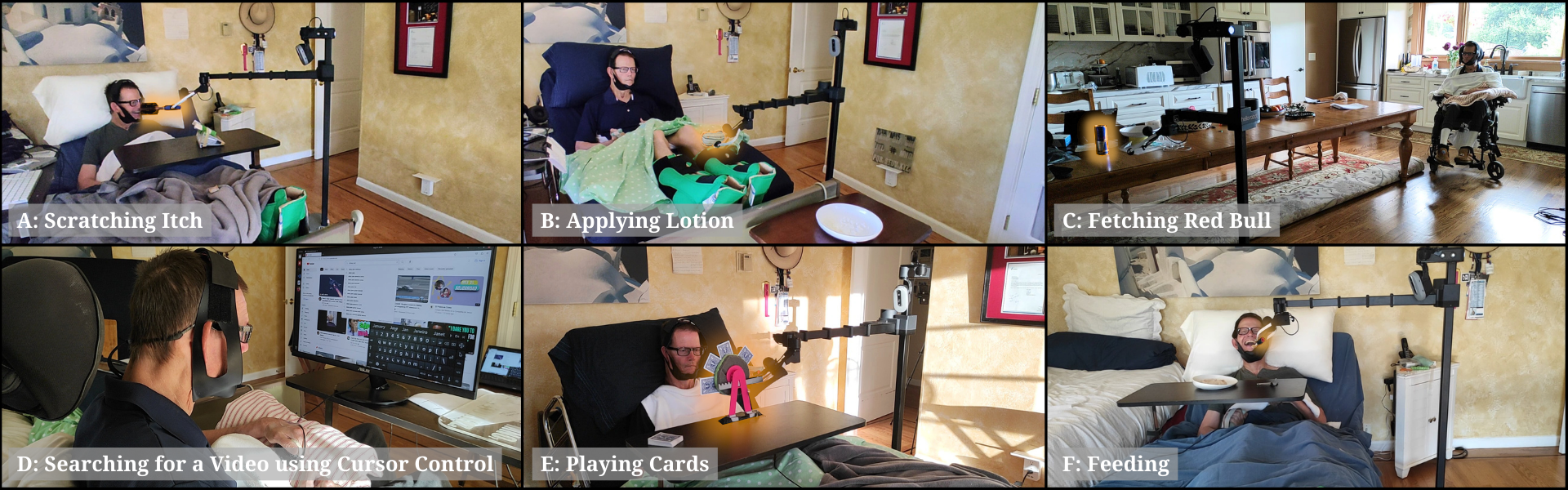}
      \vspace{-0.7cm}
      \caption{A subset of the tasks conducted during the study. Pertinent objects and tools are highlighted in orange.}
      \Description{6 images are included showing a subset of the tasks conducted during the study. In the first task image, scratching itch, the robot is holding a hair brush which is touching Henry's nose. In the second task image, applying lotion, the robot is holding a lotion application tool which is pressed on Henry's leg. In the third task image, fetching red bull, the robot is shown aligned with a red bull can. In the fourth task image, playing video using cursor control, Henry is shown using HAT to type into the YouTube search box. In the fifth task image, playing cards, the robot is pulling on a card wheel. In the sixth task image, feeding, the robot is holding a spoon with food in front of Henry's face as Henry smiles.}
      \vspace{-0.3cm}
      \label{tasks}
   \end{figure*}

\vspace{-0.1cm}
\subsection{Robot Mapping and Thresholds}
At the start of the study, we recorded Henry's max head rotations in the roll (Y), pitch (X), and yaw (Z) axes, seen in Fig.~\ref{robot_modes}. while wearing HAT. From a resting position, looking straight ahead, his max rotations were $\pm 17$ degrees in the roll axis, $\pm 59$ degrees in the pitch axis, and $\pm 53$ degrees in the yaw axis. While his pitch and yaw motion is unrestricted in relation to a non-disabled user, his roll motion is more constrained. The adaptability of the HAT interface allowed Henry to use only the pitch and yaw axes of his head motion for robot control. 

The robot receives velocity commands over WiFi from the companion laptop at 20 Hz. In drive mode, the pitch angle is mapped to the robot base's forward and backward movement, while the yaw is mapped to clockwise and counterclockwise rotation. When the robot is in arm mode, as depicted in Fig.~\ref{robot_modes}, pitch controls the height of the robot arm, and yaw manages its extension. Lastly, when in wrist mode, pitch controls the opening and closing the gripper, while yaw controls the wrist's rotational movement. Upon entering either robot control or cursor control mode, Henry must first initialize the orientation of HAT by clicking once. During initialization, the orientation of the interface is saved and used for setting four thresholds: the two minimum motion thresholds, $t_{l,f}$ and $t_{l,b}$, and the two maximum motion thresholds, $t_{h,f}$ and $t_{h,b}$, as shown in Figure~\ref{robot_modes}. In conjunction with the user's head orientation, these thresholds are used for calculating the velocity sent to each robot actuator. After testing with Henry at the start of the study, we set the minimum thresholds to $\pm 10^{\circ}$ from the initialized position, $\theta_i$, and the maximum thresholds to $\pm 35^{\circ}$ from the initialized position, $\theta_i$. These thresholds are shown along the pitch axis in Fig.~\ref{robot_modes} and work as follows. If the user's head is tilted less than $t_{l,f}$ and greater than $t_{l,b}$, in both the X (pitch) and Z (yaw) axes, the robot stops all motion. While the angle measurement along an axis is within a pair of minimum and maximum motion thresholds (e.g. $t_{l,f}$ and $t_{h,f}$), the measurement is proportionally scaled to a velocity command for the robot's actuators. Further details and a mathematical formulation for velocity scaling are provided in Appendix~\ref{velocity_scaling}.

%\change{The methodology for choosing the velocity limits, the specific limit values, and an equation for computing $k_a$ can be in found in Appendix~\ref{velocity_limits}.}

%were set at $15^{\circ}$ and $-15^{\circ}$ respectively from the initialized position, $\theta_c$. The maximum thresholds, $t_{h,f}$ and $t_{h,b}$, were set at $45^{\circ}$ and $-45^{\circ}$ respectively from the initialized position, $\theta_c$. 

% allowing Henry to commence robot motion with a smaller head movement from his resting position than before.
%allowing him to reach the top robot speed with a smaller head movement from his resting position than before. 

%The yaw (Z) axis of the IMU is only used for triggering speech recognition and not for any robot motion, allowing the user to rotate their head in this axis to look at objects in their environment.

\vspace{-0.1cm}
\subsection{Cursor Control}
To investigate the adaptability of the interface for use with different physical devices, we explored using HAT to control a computer cursor. This idea was proposed by Henry, who currently uses a computer vision-based head tracking interface that uses a webcam to track a reflective dot attached to his glasses. Prior to the start of the study, we implemented velocity control of the cursor, which uses the same equation as robot control. In this setup, tilting your head along the pitch axis moves the cursor up or down, and tilting it along the yaw axis shifts the cursor left or right. The degree of head tilt determines the cursor's speed. Based on Henry's suggestions, provided on Day 1, we developed and tested a position control version on Day 2 and Day 3, which acts more similarly to his vision-based head tracking interface. Using position control, the user's head orientation is directly mapped to a cursor position on the screen using linear interpolation, described further in Appendix~\ref{position_control}. 

\begin{figure*}[hbt!]
      \centering
      \includegraphics[width = \textwidth]{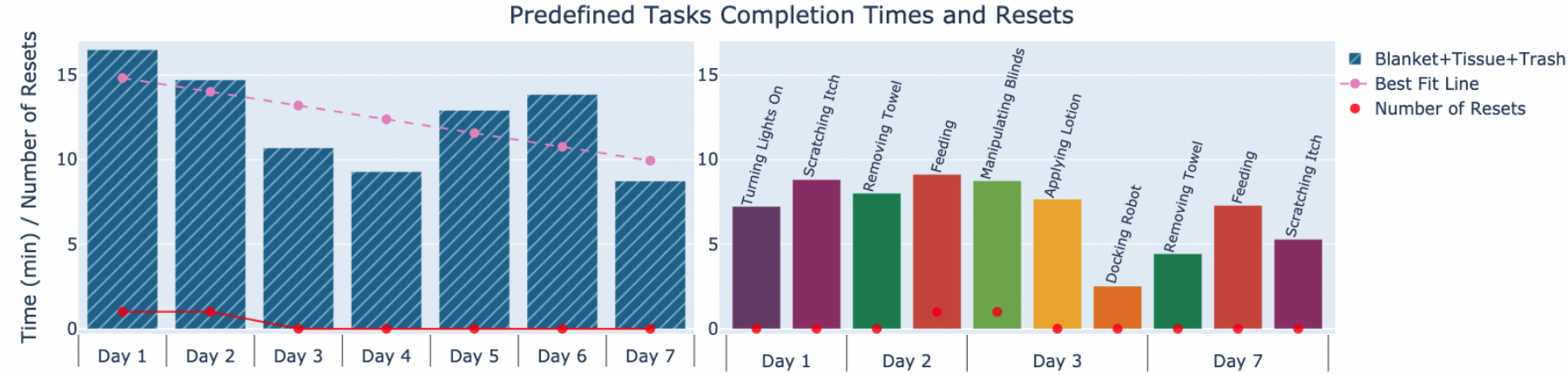}
      \vspace{-0.6cm}
      \caption{Left: Completion time and resets for the repeated blanket+Tissue+Trash task. A best fit line is plotted, showing a downward trend in completion times over the course of the study. Right: Completion time and resets for all other predefined tasks. Improvement is seen in task completion times and resets required for all repeated tasks over the course of the study.}
      \Description{On the left, the completion times and resets for the repeated Blanket+Tissue+Trash task are shown. A best fit line is also shown, showing a downward trend in task completion times for this task. On the right, are the completion times and resets for the remaining tasks. The raw data for this plot is included in Table A3.}
      \vspace{-0.4cm}
      \label{task_times}
   \end{figure*}

%Following the third day, we attempted to refine the crsor control by experimenting with filtering algorithms on the IMU data.

\vspace{-0.1cm}
\subsection{Graphical User Interface}
Previous instantiations of head-worn interfaces have relied on line-of-sight~\cite{padmanabha2023hat, wheelchairIMU, wheelchair2, wheelchair3}. In contrast, when we pair HAT with a graphical user interface (GUI), depicted in Fig.~\ref{gui}, users can control a mobile manipulator at further distances, performing tasks in other rooms, such as retrieving a drink from the kitchen to the bedroom. The GUI provides information about the current mode, how to access other modes, and notably, it features two views from fisheye cameras, one mounted at the top of the robot for navigation and one mounted at the gripper for precision alignment and grasping. 

%In the previous HAT study, all tasks were carried out with line-of-sight. Without a graphical user interface (GUI), users are limited to performing tasks only in close proximity and may not be able to command the robot to complete tasks that are farther away, such as tidying up in different rooms, fetching an object, or moving a laundry basket. Thus, to evaluate the functionality of HAT in scenarios where direct line-of-sight is not possible, we developed a GUI, depicted in Figure~\ref{gui}. 
   
\subsection{Driver Assistance}
To minimize cognitive load for grasping tasks, we propose a shared control method for HAT called Driver Assistance. When activated by the user, the system first processes input language queries that define target objects and matches these to detected objects within the environment. It then automatically aligns the robotic gripper's position with the intended object, while ensuring that the user maintains a sense of control.

\subsubsection{Modeling}
We model the Stretch robot, shown in Fig.~\ref{robot_modes}, as a nonholonomic mobile manipulator. The velocity kinematics of the robot are $\mathcal V = \boldsymbol{J}(\boldsymbol{\theta})[ v,\, \omega,\, \dot{\boldsymbol{\theta}} ]^\mathsf{T}$, where $\mathcal V\in\mathbb{R}^6$ is the twist of the end-effector, $v$ is the mobile base's forward velocity and $\omega$ rate of rotation, and $\boldsymbol{\theta} \in \mathbb{R}^3$ consists of the joint angles and joint positions of the arm. Lastly, $\boldsymbol{J}(\boldsymbol{\theta}) \in \mathbb{R}^{6\times 5}$ is the Jacobian matrix mapping between joint velocities and end-effector twist in robot base frame.
% $\boldsymbol{J}(\boldsymbol{\theta})$ can be written as the concatenation of base Jacobian and arm Jacobian.
% \begin{equation}
%     \boldsymbol{J}(\boldsymbol{\theta}) = \begin{bmatrix}\boldsymbol{J}_{\mathrm{base}}(\boldsymbol{\theta}) & \boldsymbol{J}_{\mathrm{arm}}(\boldsymbol{\theta})\end{bmatrix}.
% \end{equation}

\subsubsection{Open-Vocabulary Perception}
 To handle perception in complex household environments, we use the robot's Intel RealSense D435i RGB-D in conjunction with OWL-ViT~\cite{minderer2205simple}, an open-vocabulary object detection model that accepts natural language (text-based) queries describing the object of interest for grasping. After obtaining a bounding box from OWL-ViT, the system segments the corresponding object from the point cloud, and then locates the pose of the object in the world frame for grasping. In this study, we input the language queries, listed in Table~\ref{tab:da_queries}, for our participant based on the Driver Assistance tasks detailed in Section~\ref{study_design}. We envision that users will select their own queries via a GUI in future iterations of our system.

\subsubsection{Intent Recognition}

The perception system provides a set of potential target objects denoted as goal set $\boldsymbol{g}$. From the goal set, we must infer the most probable object $g^*\in\boldsymbol{g}$ that the user wants to grasp based on the robot's state $\boldsymbol{\Theta}$. For each target object $g\in\boldsymbol{g}$, we design the probability of the object as the intended goal with $P(g\mid \boldsymbol{\Theta}) = 1/(1+\rho d_{\mathrm{ee}}(g))$, where $\rho=5$ is an adjustable parameter and $d_{\mathrm{ee}}(g)$ is the Euclidean distance between the end-effector and the object. Thus, the most probable goal is defined as
\begin{equation}
    g^* = \argmax_{g\in \boldsymbol{g}}P(g\mid \boldsymbol{\Theta}) = \argmin_{g\in \boldsymbol{g}}d_{\mathrm{ee}}(g),
\end{equation}
which is the closest object to the end-effector. The confidence of the current inference among all goal candidates is computed as:
\begin{equation}
    C(\boldsymbol{g}) = P(g^*\mid \boldsymbol{\Theta})-\max_{g\in\boldsymbol{g}\setminus g^*}P(g\mid \boldsymbol{\Theta}),
\end{equation}
 which is the difference between the probability of the most likely target object and that of second most likely target object~\cite{jain2020probabilistic}. 
 
 %The confidence equation describes the uncertainty of the current goal inference.

%We need to infer the most probable object $g^*\in\boldsymbol{g}$ that the operator wants to grasp. The inference is based on the robot's state $\boldsymbol{\Theta}$.

% We take a simple assumption that the nearer the end-effector is to the goal, the higher is the probability of that goal.
%We simplify the inference by assuming that the closer the end-effector approaches a goal within the set, the more likely it aligns with the operator's intent.

\subsubsection{Driver Assistance Controller}

Consider a kinematic controller where the control input, $\boldsymbol{u}\in\mathbb{R}^5$, is executed on joint velocities.
% \begin{equation}
% \begin{bmatrix}
%         v& \omega & \dot{\boldsymbol{\theta}}
%     \end{bmatrix}^\mathsf{T} = \boldsymbol{u},
% \end{equation}
To share control between the user and robot, we use a control blending method $\boldsymbol{u} = \boldsymbol{u}_h + \alpha\boldsymbol{u}_a$ similar to past literature~\cite{gopinath2017human,dragan2013}, where $\boldsymbol{u}_h\in\mathbb{R}^5$ is the operator's input, $\boldsymbol{u}_a\in\mathbb{R}^5$ is the input from the robot's controller, and $\alpha$ is a weighting factor determining the extent of assistance the robot should provide. In our design, we set $\alpha=C(\boldsymbol{g})$, resulting in the robot providing little to no assistance when its confidence of the goal is low. The robot's controller is formulated as a Cartesian space proportional controller $\boldsymbol{u}_a = \boldsymbol{K}_p\boldsymbol{J}^\dagger(\boldsymbol{\theta})(\boldsymbol{x}_d-\boldsymbol{x})$, where $\boldsymbol{K}_p \in \mathbb{R}^{5\times 5}$ are the control gains, $\boldsymbol{J}^\dagger(\boldsymbol{\theta})$ is the pseudo inverse of the Jacobian matrix, $\boldsymbol{x}_d$ is the desired end-effector configuration which is the pose of $g^*$, and $\boldsymbol{x}$ is the current end-effector configuration. To ensure a user retains the feeling of control of the robot, we constrain the robot's assistance on the joint that the user is controlling by multiplying a factor $\rho_c=0.2$ on the corresponding row of $\boldsymbol{u}_r$. We additionally prevent the robot from controlling the arm extension, base translation, and wrist rotation by setting $\boldsymbol{K}_p = \operatorname{diag}(0,1,1,0,0)$. By constraining these specific joints, the robot’s motion only facilitates alignment with the object, allowing the user to remain in control of how the robot approaches and grasps the object with the arm.

%Additionally, we give the user full control on arm extension and freeze forward speed and wrist rotation. This is achieved by setting the control gain of corresponding joints in $\boldsymbol{K}_p$. The final control gain matrix is $\boldsymbol{K}_p = \operatorname{diag}(0,1,1,0,0)$.

\vspace{-0.1cm}
\section{Study Design}
\label{study_design}
A full, comprehensive study schedule can be found in Fig.~\ref{calendar}. The IVs are the task and the use of driver assistance and the DVs are task completion time and perception of the system. Henry completed a number of predefined tasks with line-of-sight of the robot: blanket+tissue+trash, scratching an itch, feeding, and applying lotion. Detailed task descriptions and completion criteria can be found in Appendix~\ref{tasksetup}. He additionally completed several tasks without line-of-sight using the GUI. These tasks include turning the light on, manipulating blinds, removing soiled towel, and docking robot. The predefined tasks were primarily chosen based on tasks that Henry had previously completed in past assistive robotics studies~\cite{robotsforhumanity, surrogates, tasksaroundhead, tasks, nguyenincreasing}. Each day of the week-long study, he additionally performed a consistent challenging task, blanket+tissue+trash, shown in Fig.~\ref{fig:teaser}, to track performance with increasing experience over time. Except for this task, all predetermined tasks were executed once in the initial three days of the study. On the final day, we also repeated three of the predefined tasks (scratching itch, removing soiled towel, and feeding) to further assess improvement. Some tasks involved custom tools, like a soft silicone spoon with an adapted handle for improved robot grip, as shown in Fig.~\ref{tasks}F. Details about tools and modifications of the home environment for each task can be found in Appendix~\ref{tasksetup}.
 
On the 5th day of the study, we had Henry do 3 within line-of-sight tasks to systematically evaluate our Driver Assistance method. The tasks were fetching Red Bull, cleaning up cups, and removing soiled towel. The tasks were completed with and without Driver Assistance, with the starting order alternated for each consecutive task. Two of the tasks were purposefully designed to be challenging, to explore the potential benefits of Driver Assistance. Specifically, for the fetching Red Bull task shown in Fig.~\ref{tasks}C, we placed the drink can about 10 feet from Henry, making it challenging for depth perception. The cleaning up cups task also presented a challenge as it demanded manipulating two cups in sequence.

Throughout the study, we encouraged Henry to suggest any tasks he may want to attempt with the robot. On the 2nd day of the study, Henry used the robot for playing cards with his wife using a card wheel, shown in Fig.~\ref{tasks}E, that can be rotated by the robot. On the 6th day of the study, we conducted an open day of testing, allowing Henry to dictate the tasks he wanted to try for an entire day. Henry did 4 tasks (opening drawer+fetching tape, grasping+fetching bottle, grasping towel+wiping table, and opening fridge), some with driver assistance and some out of line-of-sight.  

After each task and at the end of each day, Henry responded to multiple 7-point Likert and NASA TLX workload items about robot control listed in Tables~\ref{tab:sup1}-\ref{tab:sup4}. He was additionally asked open-ended questions, included for reference in Appendix~\ref{questionnaires}.

%After completion of each task and at the end of each day, Henry was asked multiple 7-point Likert Items related to ease of use, errors, error recovery, time, and preference for the HAT interface over a web interface. Henry was asked additional Likert items if the task involved cursor control or the GUI. After each Driver Assistance task, Henry was asked to fill out the NASA TLX workload 7-point scale and additional Likert items related to his perception of control, whether the robot did what he wanted it to do, efficiency, reduction in frustration, and preference for Driver Assistance. For the open tasks, Henry was asked the pertinent Likert items from the aforementioned. At the end of the day, Henry was also asked to fill the NASA TLX scale and open-ended questions. All scales and questions can be found in the Appendix. 

\vspace{-0.1cm}
\section{Results and Discussion}
For robot control tasks, task completion times, number of resets, and responses for Likert items and the NASA TLX scale can be found in Tables~\ref{tab:sup1}-\ref{tab:sup4}. 6 tasks are shown visually in Fig~\ref{tasks}. Task videos are included on the project website.

\vspace{-0.1cm}
\subsection{Metrics}
\subsubsection{Task Completion and Efficiency}
Henry attempted and completed a total of 17 predefined tasks, whose completion times are shown in Fig.~\ref{task_times}. Times were calculated based on the conditions detailed in Appendix~\ref{tasksetup} and were modified to account for any technical issues. Through the course of the study, Henry became increasingly more efficient, showing a clear improvement in usage of the interface over time. This was observed in the recurring task, blanket+tissue+trash task, where there is a noticeable downward trend in completion times as seen on the left side of Fig.~\ref{task_times}. Specifically, from day 1 to day 7, there was a 47\% (7 m 46 s) decrease in time. With all 3 repeated predefined tasks, scratching itch, removing soiled towel, and feeding, we see clear, positive improvements from the start to the end of the study with a 40\% (3 m 32 s), 45\% (3 m 35 s), and 20\% (1 m 50 s) reduction respectively in task times.

Over the week long study, Henry's perception of the interface's ability to enable task completion in a reasonable amount of time improved with him answering 4 (Neutral) and 2 (Disagree) on Days 1 and 2 to 6 (Agree) on Day 7. When asked about the efficiency of the interface in comparison to a direct teleoperation web interface without autonomy, Henry said ``HAT is slightly more efficient [than a web interface] once you have mastered it. Certain specific tasks, however, are definitely more efficient with HAT. HAT in all cases is markedly better, and significantly more efficient and effective at 1) face to face things like playing simple games, or whenever a computer in your face is a problem like [during] blanket manipulation, 2) fine motor control of [the] gripper, and 3) [tasks that] use Assisted Grasping (Driver Assistance)--which is awesome and should be added to every Stretch interface.'' At numerous points of the study, Henry mentioned that the ``analog'' (continuous) control of the robot (i.e. degree of head tilt mapped to velocity of actuators) was useful, leading to higher efficiency and more precise control in comparison to direct teleoperation web interfaces. 

%Median results from Likert items asked after each predefined task are shown in Table~\ref{tab:predefined_tasks}. Across all predefined tasks, Henry's median response was 4 (Neutral) for the Likert item about the task taking a reasonable amount of time.

\begin{figure}[t!]
      \centering
      \includegraphics[width = \columnwidth]{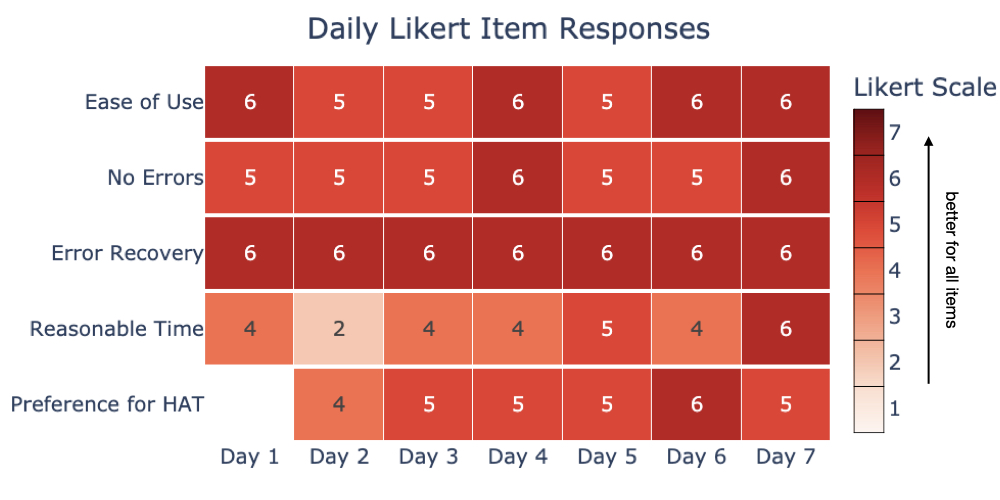}
      \vspace{-0.6cm}
      \caption{7-point Likert item responses for Ease of Use, No Errors, Error Recovery, Reasonable Time, and Preference for each day of the study. For all items, a higher rating is better.}
      \Description{The 7-point Likert item responses for Ease of Use, No Errors, Error Recovery, Reasonable Time, and Preference for HAT, for each day of the study. For all items, a higher rating is better. The raw data for this plot is in Table~\ref{sup4}.}
      \vspace{-0.45cm}
      \label{likert_daily}
   \end{figure}

\subsubsection{Errors and Recovery}

Our findings show that with minimal practice Henry was able to limit errors during tasks and consistently recover from any errors that occurred, two essential characteristics for any proposed teleoperation interface for mobile manipulators in a home environment. As seen in Table~\ref{tab:predefined_tasks}, across all predefined tasks, Henry's median response was Agree (6) to the Likert item ``I was able to complete the previous task without any errors using the control interface''. Additionally, as seen in the daily responses to the same Likert item in Fig.~\ref{likert_daily}, Henry replied Somewhat Agree (5) or Agree (6) on all days. Henry completed 13 out of 17 predefined tasks without any resets (critical errors that required resetting of the environment or robot by a researcher). The number of resets per predefined task is shown in Fig.~\ref{task_times}. For the blanket+tissue+trash task, Henry required a reset on the first two days of attempting the task. More details on resets can be found in Appendix~\ref{resets}. After Day 3, no resets were needed as Henry succeeded in instinctively putting the robot back into idle mode and re-initializing the HAT when he faced any inadvertent robot motion. As seen in the median responses across all predefined tasks (Table~\ref{tab:predefined_tasks}) and daily responses (Fig.~\ref{likert_daily}), he responded Agree (6) to the Likert item ``If there were errors, the control interface allowed easy recovery''.

   \begin{figure}[t!]
      \centering
      \includegraphics[width = \columnwidth]{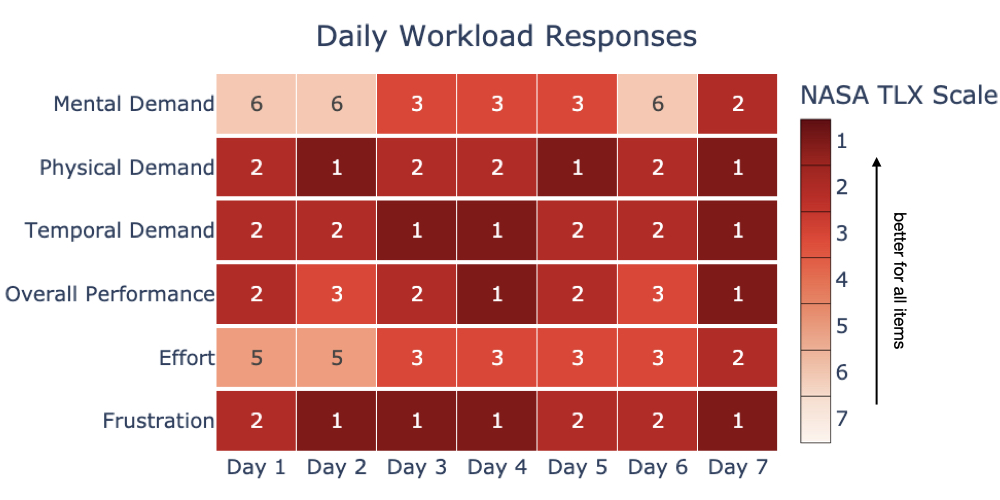}
      \vspace{-0.6cm}
      \caption{7-point NASA TLX responses for each day of the study. For all items, a lower rating is better.}
      \Description{The 7-point NASA TLX responses for each day of the study. For all items, a lower rating is better. The raw data for this plot is in Table~\ref{sup4}.}
      \vspace{-0.25cm}
      \label{nasatlx_daily}
   \end{figure}

\begin{table}
  \caption{Predefined Tasks Likert Responses}
  \vspace{-0.3cm}
  \label{tab:predefined_tasks}
\begin{tabular}{cccccc}
\toprule
Likert Item        &  median &  interquartile range &  min &  max \\
\midrule
Ease of Use        &       6 &    0 &    3 &    7 \\
No Errors             &       6 &    1 &    3 &    6 \\
Error Recovery     &       6 &    0 &    5 &    7 \\
Reasonable Time    &       4 &    1 &    2 &    5 \\
Preference for HAT &       5 &    0 &    4 &    6 \\
\bottomrule
\end{tabular}
\vspace{-0.35cm}
\end{table}

%$Other than critical errors, we identified a few more common errors. Henry faced  common errors such as overshooting and accidentally tilting his head the wrong way. Additionally, Henry faced more inadvertent robot motions which were exacerbated for three reasons in comparison to our past study: 1. Thresholds for motion sometimes changed over time due to more IMU drift from magnetic interference in the home environment 2. Henry used the yaw axis instead of roll for control of the robot, which limited his ability to look around the room while controlling the robot. 3. Henry experiences twitches and spasms. Despite Henry finding HAT to be more prone to erratic movements than a web interface due to the above reasons, 

\subsubsection{Ease of Use}
As seen in Table~\ref{tab:predefined_tasks}, across all predefined tasks, Henry's median response was Agree (6) to the Likert item ``The control interface was easy to use.'' Specifically, Henry gave a 7 (Strongly Agree) ranking to 3 tasks: Playing Cards, Blanket+Tissue+Trash, and Manipulating Blinds. This was similarly the case for his daily responses; as seen in Fig.~\ref{likert_daily}, Henry replied either Somewhat Agree (5) or Agree (6). As discussed in the Task Completion and Efficiency section, Henry found the interface well suited for any precise tasks due to having continuous control and for any tasks around his body due to not having a monitor directly in front of him. At the end of the study, Henry specified there are certain tasks that he would use HAT and the robot to do instead of relying on a caregiver. He said, ``Definitely scratching itches. I would be happy to have it stand next to me all day, ready to do that or hold a towel to my mouth. Also, feeding me soft foods, operating the blinds and doing odd jobs (`freestyling') around the room would be helpful.'' Starting on the second day of the study, we additionally asked the Likert item, ``I prefer to use HAT over a web interface overall'' assuming direct teleoperation and no robot autonomy for both interfaces. As seen in Fig.~\ref{likert_daily} and Table~\ref{tab:predefined_tasks}, Henry either answered Neutral (4) or higher for all tasks, with a median response of Somewhat Agree (5). When asked if the HAT interface is intuitive to use, Henry said, ``It takes a bit of practice, but all controls become reflexive after a few days.''

   \begin{figure*}[hbt!]
      \centering
      \includegraphics[width = \textwidth]{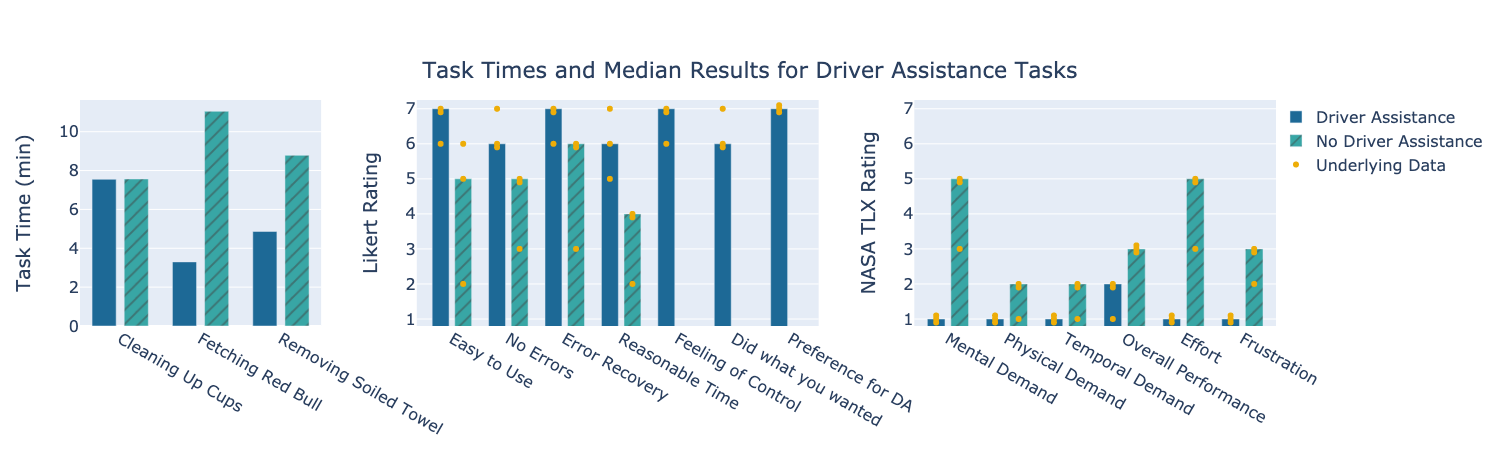}
      \vspace{-0.7cm}
      \caption{Left: Comparison of task times for the 3 tasks completed with and without Driver Assistance. Middle: Median Likert items  across all 3 tasks with and without Driver Assistance. For all Likert items, a higher rating is better. 
      Right:  Median NASA TLX responses across all 3 tasks with and without Driver Assistance. For all NASA TLX items, a lower rating is better. }
      \Description{The left side of this figure shows task completion time comparisons for each of the three tasks conducted with and without Driver Assistance. The middle shows all Likert Item responses for the tasks with and without Driver Assistance. Lastly, the right side shows all NASA TLX responses for the tasks with and without Driver Assistance. Raw data for this figure can be found in Table~\ref{sup3}.}
      \vspace{-0.4cm}
      \label{DA_results}
   \end{figure*}

\subsubsection{Learning Curve}
By comparing task times and errors for the daily blanket+tissue+trash task, we can more thoroughly explore the interface learning curve. After two days, Henry was able to complete the task without any resets required, showing familiarity with the interface. From the second to third day of the study, his perception of errors also improved with a jump in response from 3 (Somewhat Disagree) to 6 (Agree) for the Likert item on no errors. Despite some outliers, Henry continued to improve his task time over the course of the study as shown by the best fit line in Fig.~\ref{task_times}. Henry commented on the learning curve saying, ``It actually had a fairly steep learning curve, although not wide. An easy way to reduce the learning needed would be to produce a `Quick Start' card providing, on one side, all the information you need to get started, including what each head movement does. After a few days, the card will be unnecessary. It does get easier every day for a while.''

\subsubsection{Workload}
Daily workload measures using a 7 point NASA TLX Scale are shown in Fig.~\ref{nasatlx_daily}. On the first two days, Henry reported higher values of mental demand (6) and effort (5). He specified this was due to the clicking patterns (which we changed after day 2) and due to lack of familiarity with the interface. On day 6, the open day, there was another spike in mental demand due to some of the challenging tasks (opening drawer+tape fetching and opening fridge) that Henry did with the interface and robot. Throughout the study, Henry had strong ratings for HAT in physical and temporal demand, overall performance, and frustration. Henry gave his strongest workload ratings for HAT on the last day with a 1 (Low) for physical demand, temporal demand, and frustration, a 1 (Successful) for overall performance, and a 2 for mental demand and effort. While evaluation with more individuals with impairments is needed, these strong workload measures highlight the potential for HAT for individuals with motor impairments who retain some degree of head and neck mobility.

\vspace{-0.1cm}
\subsection{Effect of Driver Assistance}
We find that Driver Assistance led to clear improvements in task times, Likert responses, and workload measures while still preserving user control of the robot in the home. As seen in Fig.~\ref{DA_results}, for cleaning up cups, fetching Red Bull, and removing soiled towel, Driver Assistance led to a reduction in task times by 1 s, 7 m 45 s (70\%), and 3 m 55 s (46\%) respectively. Despite similar task times for the cleaning up cups task, while not using Driver Assistance, Henry required a reset of the environment for knocking over one of the cups while trying to align the robot. For the fetching Red Bull task, the automatic alignment was especially essential as the can was far away from Henry, making it extremely challenging to perceive depth to align the gripper with the can. 

After testing Driver Assistance for the first time, Henry expressed his fondness for the system saying, ``that's better than anything I have tried for grasping.'' As seen in Fig.~\ref{DA_results}, Henry reported improvements in ease of use, errors, error recovery, and completion time for all tasks. Similarly, in Fig.~\ref{DA_results}, Henry reported a substantial reduction of 4 points in both mental demand and effort and improvements in physical/temporal demand, overall performance, and frustration. In contrast to prior work in shared control where users have reported reduced feeling of control with increasing robot autonomy~\cite{javdani2018,bhattacharjee2020}, Henry did not feel this for HAT. For the Driver Assistance tasks, as seen in Fig.~\ref{DA_results}, Henry agreed (6) or strongly agreed (7) with the Likert items related to feeling in control, having the robot do what he wanted, and preference for the assistance over full teleoperation. All raw data relating to Driver Assistance are in Table~\ref{tab:sup3}. 

Further testing will be necessary to assess the advantages of driver assistance in improving robotic manipulation within a natural home environment for a broader population of individuals facing significant motor impairments. Nevertheless, these preliminary findings show that incorporating open-vocabulary object detection perception models alongside shared control for alignment during grasping tasks may play a vital role in alleviating the workload involved in robotic teleoperation within household settings.

\vspace{-0.1cm}
\subsection{Out of Line-of-Sight Tasks}

Henry completed 5 predefined tasks (turning the lights on, manipulating blinds, docking robot, and removing soiled towel twice) out of line-of-sight using HAT coupled with a GUI displaying the robot's camera feeds. On the open day, he completed two more tasks, opening drawer+fetching tape and opening fridge, using HAT and the GUI. Across all 7 tasks, when asked the Likert Item, ``The control interface was harder to use without line-of-sight than with line-of-sight,'' Henry responded with a median response of 1 (Strongly Disagree) with an interquartile range (IQR) of 1, showing that the experience of using HAT with the GUI was similar to using it without. On Day 3, after completing all the predefined tasks at least once, Henry said controlling the robot using the GUI is ``much easier [than without] since [the] yaw [axis] control makes it difficult to look around.'' Although, at times, Henry found the GUI camera views useful, he mentioned on numerous occasions that he preferred a screen-free experience when performing tasks involving his face and body.

%In one task, docking, we also evaluated how Henry switched between using HAT for cursor control and for robot control. At the start, Henry had to first play a video on YouTube on his computer before docking the robot. After docking, he had to return to cursor control to pause the video. Henry responded 4 (Neutral) to the Likert item: "The control interface was intuitive to use with both the computer and the robot for the previous task." 

\vspace{-0.1cm}
\subsection{Control of Multiple Devices}
On multiple days of the study, Henry evaluated HAT as an interface to control the cursor on his computer. Two tasks that Henry completed were typing sentences in a notepad and searching for a video on YouTube as seen in Fig.~\ref{tasks}D. On Day 3, for a quickly implemented position-based cursor control, the system worked well with Henry saying it was ``pretty close'' in comparison to his current head tracking interface though ``about 60\% as fast as [the] head tracker because [the] cursor jumped around''. One notable advantage, as highlighted by Henry, was that HAT required less frequent recalibration compared to his existing head tracking system. Our changes after Day 3 concentrated on mitigating cursor jumps caused by noise in the IMU-estimated angles. Testing of cursor control with the HAT showed that the interface is viable for control of other physical devices. Much like a web interface, one can envision HAT seamlessly enabling concurrent tasks such as controlling a robot, composing emails on a computer, and changing channels on a TV, all without the need for caregiver assistance.

\vspace{-0.1cm}
\subsection{Lessons Learned}
\subsubsection{Viability of HAT}
In conjunction with our previous study on this interface which involved 18 participants of primarily low expertise with controlling robots, the results of this case study show that HAT is a strong option as a direct teleoperation interface for mobile manipulators as it performs strongly in terms of task efficiency, errors, ease of use, learning curve, and workload. These results are further supported by our participant’s preference for HAT in comparison to similar direct teleoperation web interfaces he has used in the past. We expect these results to generalize to a broad range of individuals with motor impairments, though head and neck motor function along at least 2 axes must be retained for the interface to be effective. Since extensive testing was only conducted with a non-speaking individual with quadriplegia due to a brainstem stroke, future testing with additional individuals with motor impairments is crucial to fully validate these findings.

\subsubsection{Customizability}
This work highlights the importance of personalizing assistive interfaces to individual needs. In the case of Henry, utilizing prior versions of HAT would have been infeasible without switching from roll to yaw axis control and from speech to clicker for mode switching. Furthermore, without the ability to adjust mode switching clicking patterns to suit his specific needs and condition, Henry would have also had significantly higher mental demand while using the interface. Lastly, the ability to modify the physical design of HAT holds significance, allowing users to incorporate their comfort and fashion preferences into the interface. 

%To facilitate customizability for HAT, a settings menu in the GUI could enable the individual or their caregiver to make adjustments as necessary. 

\subsubsection{Combining Interfaces}
Wearable interfaces, such as HAT, offer distinct advantages when compared to standard interfaces for robotic teleoperation. First, wearable interfaces embedded in clothes are readily accessible when needed and inconspicuous when not in use. Such interfaces can offer a more direct and intuitive means of controlling mobile manipulators. Additionally, the absence of a screen or input device (head tracking, sip and puff, etc.) directly in front of the user's face allows for increased flexibility when performing tasks around the body and enhances situational awareness. On the other hand, web-based interfaces come with their own set of benefits, including a greater number of inputs through on-screen buttons and the capability to display camera feeds. They also simplify the activation of autonomous routines, allowing for completion of repetitive tasks such as navigating between rooms or turning on/off the lights with a single click. At the conclusion of the study, Henry specified that the largest limitation of HAT was the lack of autonomous routines other than Driver Assistance. His feedback was that we could ``use a web interface to `program' the automated function and either HAT or voice controls to execute it''.

Combining wearable technologies like HAT with a screen-based interface for triggering semi-autonomous or autonomous robot actions could present a robust interface for many individuals with motor impairments. It's conceivable that Henry might employ HAT to fully teleoperate the robot to assist him at times of rest when he doesn't have his computer screen in front of him. Many of his needs at this time such as wiping his face, itch scratching, or moving his blanket may be benefited by limited autonomy, as this would allow him to exercise preference. In contrast, at times when his caregiver has set up his computer in front of him, he could use HAT with the screen-based GUI. With the presented cursor control functionality, he could use the GUI to trigger a variety of autonomous actions that pair with direct teleoperation using HAT. To reduce a view-obstructing screen, future research could explore the use of augmented reality or the display of information via a projected screen at a distance from the user's body. 

\subsubsection{Adoption in the Home}
Intuitive teleoperation of assistive mobile manipulators in the home is approaching a technologically feasible state with a wearable interface like HAT. For wider adoption, a few more development steps are needed for safety and robustness. 

Henry occasionally encountered IMU drift, likely due to magnetic interference, sometimes leading to unintended robot movements as a result of shifting thresholds. We found that this drift was notably pronounced when Henry was seated in his wheelchair and his legs were involuntarily shaking, causing his head to also shake. Under the existing system setup, Henry had no means of realizing drift was occurring except when it resulted in unintentional robot motions. Many IMUs, including the one utilized, provide calibration status indicators which could be used to require the user to reinitialize HAT when the confidence in the fusion estimate is low. 

During the human study, Henry also faced inadvertent robot motions due to spasming and looking around the room without putting the robot into idle mode. Another potential safety feature is an `anomaly detection' classifier~\cite{nassif2021machine, park2016multimodal} to distinguish the aforementioned head motions from intended head tilting movements for robot control. Given that head tilting motions with HAT tend to be notably slower than other head movements, this classifier could predict intent using head accelerations captured by the IMU and could automatically switch the robot to the idle mode upon detection of a non-head tilting motion.

Lastly, a safety module which forecasts potential collisions based on depth images and robot velocities, could enhance suitability for home use. This functionality would enable the robot to automatically decrease actuator velocities while in close proximity to the body, increasing safety and reducing mental workload.

\section{Acknowledgments} 
We thank Henry and Jane Evans for graciously welcoming us into their home for conducting the human study. We thank Hello Robot for lending a robot for the study and supporting development. Fig.~\ref{robot_modes} was created with BioRender.com.
%We would like to thank Qin Wang, Daphne Han, Jashkumar Diyora, Kriti Kacker, Hamza Khalid, and Liang-Jung Chen for their contributions to earlier versions of HAT, which served as inspiration for the ongoing phases of this project. 

\bibliographystyle{ACM-Reference-Format}
\balance
\bibliography{sample-base}

\clearpage

\appendix
\setcounter{table}{0}
\renewcommand{\thetable}{A\arabic{table}}
\renewcommand{\thefigure}{A\arabic{figure}}
\setcounter{figure}{0}

\section{Appendices}

\subsection{Device Electronics}
\label{device_electronics}
\begin{figure}[H]
      \centering
      \includegraphics[width = \columnwidth]{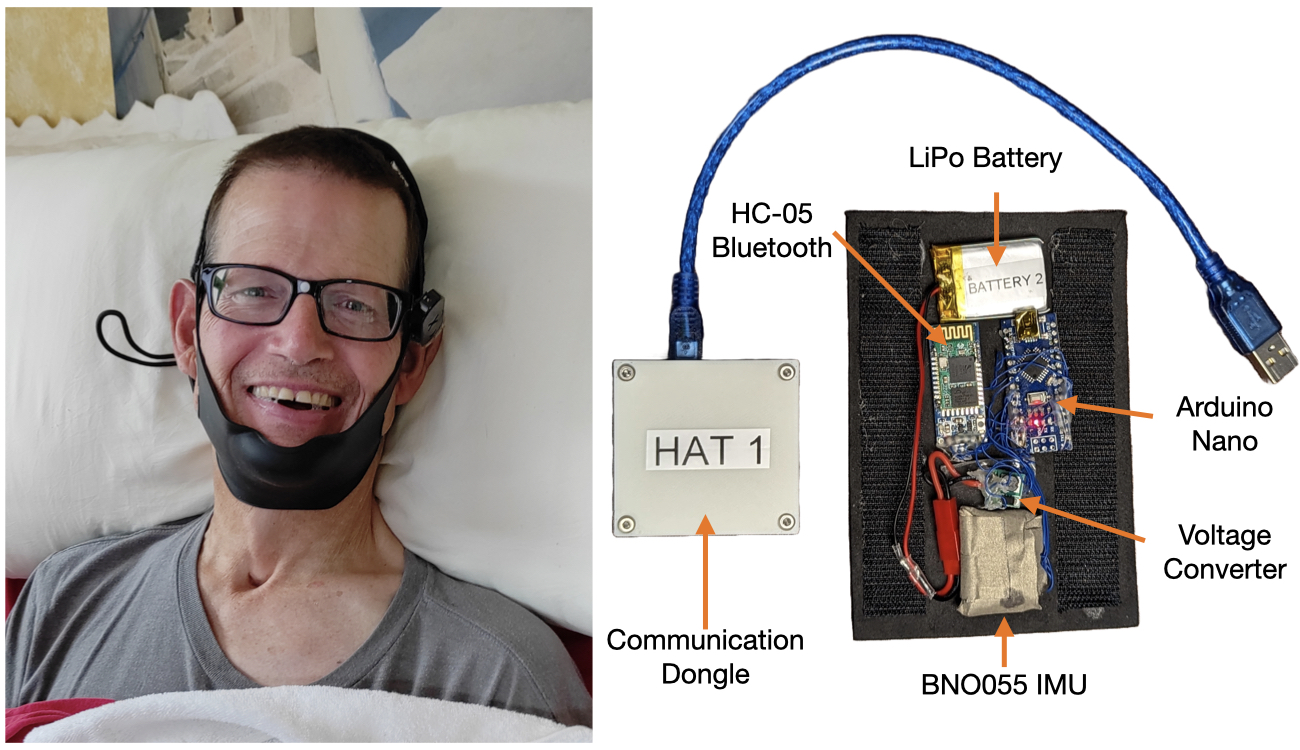}
      \vspace{-0.4cm}
      \caption{Left: Henry wearing the chin strap with the HAT interface embedded. Right: The HAT Bluetooth communication dongle and foam insert with all electronics labelled.}
      \Description{On the left, Henry is shown wearing the chin strap. On the right, is the foam insert with all the electronics labelled and the Bluetooth communication dongle. }
      \vspace{-0.4cm}
      \label{device_design}
   \end{figure}

The electronics in HAT include an Arduino Nano microcontroller, HC-05 Bluetooth module, Bosch BNO055 Inertial Measurement Unit (IMU), 5V to 3.7V voltage converter, and 400 mAh LiPo battery. These are shown in Fig.~\ref{device_design}. The Bluetooth communication dongle, which is plugged into the companion laptop, consists of an Arduino Nano and HC-05 Bluetooth module.

\subsection{Iterative Design for Mode Switching}
\label{appendix_mode_switching}
The initial clicking patterns used were as follows. While in idle mode, Henry can switch to the robot control mode using a triple click and to cursor control mode using a double click. Regardless of the current mode, he can return to the idle mode by performing two or more clicks. In the robot control mode, with a single click, Henry has the capability to cycle between driver, arm, and wrist mode. Additionally, he can hold down the clicker to activate or deactivate Driver Assistance while in robot control mode. 
After the initial two days of testing, Henry suggested, for safety, that the robot should transition into idle mode when he holds down the clicker because he often clenches his hand during spasms, coughing, or instinctively when the robot performs unintended actions. As a result, we modified the transition to idle mode to require holding down the clicker. Furthermore, we reduced the number of triple clicks required as we recognized they were more challenging for Henry to execute than single and double clicks. On day 6, after Henry's suggestion, we increased the click requirement for entering cursor control and activating Driver Assistance from 2+ clicks to 3 clicks to avoid inadvertent mode changes. These changes are shown in Table~\ref{tab:clicks}.

\begin{table}[H]
  \caption{Clicking Patterns}
  \label{tab:clicks}
  \begin{tabular}{cccc}
    \toprule
     mode switch & Day 1 & Day 3 & Day 6 \\ 
    \midrule
     idle $\rightarrow$ robot control & 3 & 1 &  1\\ 
     idle $\rightarrow$ cursor control & 2 & 2+ & 3 \\
     any mode $\rightarrow$ idle & 3+ & hold & hold \\
    drive $\rightarrow$ arm $\rightarrow$ wrist $\rightarrow$ drive  & 1 & 1 & 1\\
    Driver Assistance on $\leftrightarrow$ off & hold & 2+ & 3\\
    \bottomrule
\end{tabular}
\end{table}

\subsection{Robot Velocity Scaling}
\label{velocity_scaling}
For a single axis, the angle measurement is proportionally scaled to a velocity command for the corresponding robot actuators according to the following equation:
\begin{equation}
   V(\theta, \theta_i, a) = 
   \begin{cases} 
   -v_{a,max} & \text{if } t_{h,b} > \theta 
   \\
   k_a(\theta - t_{l,b}) &
   \text{if }  t_{l,b}  \geq  \theta \geq t_{h,b}
   \\
   0 & \text{if } t_{l,b}  < \theta < t_{l,f}
   \\
   k_a(\theta - t_{l,f}) &
   \text{if }  t_{l,f} \leq  \theta \leq  t_{h,f}
   \\
   v_{a,max} & \text{if } t_{h,f} < \theta 
  \end{cases}
\end{equation}
where $V(\theta, \theta_i, a)$ is the velocity command sent to actuator $a$, $\theta$ is the angle of the user's head, $\theta_i$ is the initial angle of the user's head during initialization, $k_a$ is the proportional constant for actuator $a$, $v_{a,max}$ is the maximum velocity limit for actuator $a$, $t_{l,f} = 10^{\circ} + \theta_i$, $t_{h,f} = 35^{\circ} + \theta_i$, $t_{l,b} = -10^{\circ} + \theta_i$, and $t_{h,b} = -35^{\circ} + \theta_i$. 

The base max translation and rotation speeds were set to $0.3$ m/s and $0.3$ rad/s respectively. The arm max lift and extension speeds were set to $0.26$ m/s and $0.13$ m/s respectively. The wrist max rotation speed was set to $0.3$ rad/s. The gripper max speed was set to $2$ rad/s. The proportional constant, $k$, for each actuator can be computed using $k_a = \frac{v_{a,max}}{t_{h,f}-t_{l,f}}$. The max velocity limits were initially set through testing with lab members prior to the study. They were marginally adjusted during the instructional session based on Henry's feedback.

\subsection{Position Control for Cursor}
\label{position_control}
Using position control, the user's head orientation is directly mapped to a cursor position on the screen using linear interpolation, as specified by the following equation: 
\begin{equation}
p(\theta, \theta_i) = p_{l} + (\theta - t_{h,b}) \frac{p_{h} - p_{l}}{t_{h,f}-t_{h,b}}
\end{equation}
where $p(\theta, \theta_i)$ is the position command sent to the computer cursor along a single axis (vertical or horizontal) of the computer screen, $\theta$ is the angle of the user's head, $\theta_i$ is the initial angle of the user's head during initialization, $p_h$ and $p_l$ are the maximum and minimum positions of the cursor along the axis of the screen, $t_{h,f} = 12^{\circ} + \theta_i$ and $t_{h,b} = -12^{\circ} + \theta_i$.

\subsection{Resets}
\label{resets}
The two resets required were: 1) Researcher stopped the task as the robot base was stuck under Henry's bed. 2) Researcher emergency stopped the robot after Henry became nervous and froze when the robot was close to colliding with him. For manipulating blinds on Day 3, we reset the task after a first attempt where Henry successfully grasped onto the blind cord and pulled it down, but the cord slipped out of the gripper. For feeding on Day 3, we reset the task after an emergency stop after Henry became nervous and froze when the robot was close to colliding with him.

\begin{table}[H]
\centering
\caption{Text Queries for OWL-ViT for Driver Assistance Tasks}
\label{tab:da_queries}
\begin{tabular}{cc}
\toprule
Task                  & Text Queries \\ \midrule
Cup Cleanup           & ``cup'', ``tumbler'' \\
Red Bull Fetching      & ``Red Bull'', ``can''\\
Removing Soiled Towel & ``cloth'', ``towel'' \\ \bottomrule
\end{tabular}
\end{table}

\begin{figure}[H]
      \centering
      \includegraphics[width = \columnwidth]{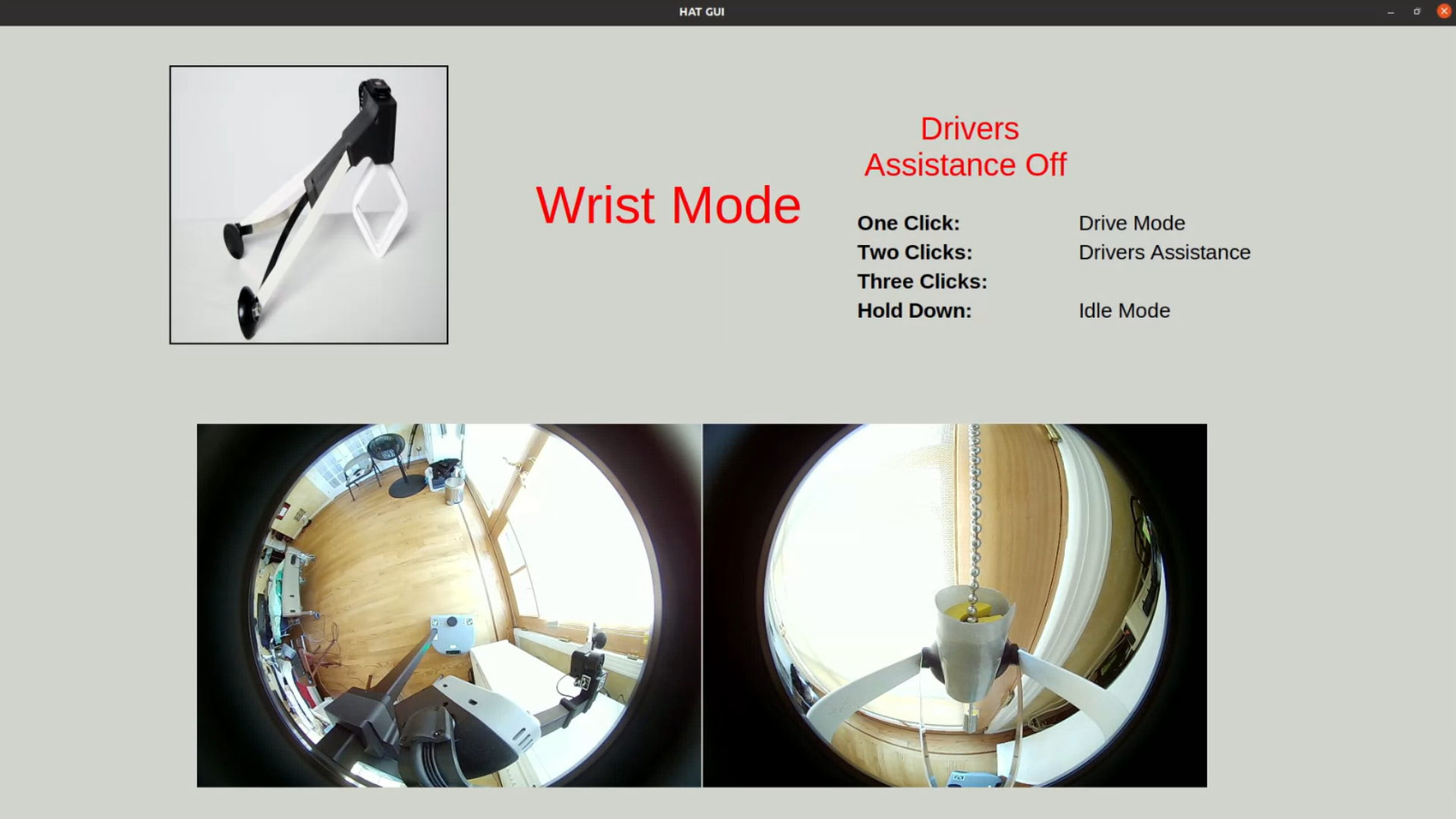}
      \vspace{-0.4cm}
      \caption{The Graphical User Interface (GUI) developed for the HAT interface.}
      \Description{The Graphical User Interface (GUI) developed for the HAT interface. The two camera feeds are shown.}
      \vspace{-0.4cm}
      \label{gui}
   \end{figure}

%\pagenumbering{arabic} 

%\setcounter{figure}{0}
%\renewcommand{\figurename}{Fig.}
%\renewcommand{\thefigure}{S\arabic{figure}}
%\renewcommand{\theHfigure}{S\arabic{figure}}
%\renewcommand{\thetable}{S\arabic{table}}
%\renewcommand{\theHtable}{S\arabic{table}}
%\renewcommand{\thesection}{S\arabic{section}}
%\renewcommand{\thesubsection}{S\arabic{subsection}}

%\setcounter{section}{0} % Reset the section counter to 0
%\renewcommand{\thesection}{S\arabic{section}} % Redefine the section counter format
%\setcounter{subsection}{0} % Reset the subsection counter to 0
%\renewcommand{\thesubsection}{S\arabic{section}.\arabic{subsection}} % Redefine the subsection counter format

%\label{sup}

\subsection{Questionnaires}
\label{questionnaires}
Participant responses to the below questionnaires were examined by researchers to identify any trends or insights. Some responses were used as quotes in the main text.
\subsubsection{Demographics and Pre-study Questions}
These were asked on the first day of the study prior to introducing the HAT or its capabilities to the participant. 
\begin{enumerate}
    \item What is your age?
    \item What is your gender?
    \item What is your ethnicity?
    \item Do you have a condition or disability?
    \item If you said yes to the previous question, what would you classify your condition or disability as?
    \item How familiar are you with using robots (where 1 is no experience and 5 is expert)? 
    \item How often do you use assistive control interfaces? Which one(s) do you use? 
    \item What is your general impression of HAT before using it? What do you think the pros and cons will be in comparison to interfaces you have used in the past?  
    \item What kind of tasks do you foresee using HAT for the most?
\end{enumerate} 
\subsubsection{Day 1 Questions} 
These were asked at the end of the first day of the study once all the tasks for the day were completed. The participant was required to finish answering these before beginning tasks on the next day to ensure that their answers were relevant to that day only. 
\begin{enumerate}
    \item What is your initial impression of HAT?
    \item What was the biggest challenge so far with using HAT? \item What are the pros and cons of HAT so far?
    \item Is there something that you would improve about the interface?  
    \item In what ways was HAT effective for the tasks that you have performed so far?
\end{enumerate} 
\subsubsection{Day 2 Questions} 
These were asked at the end of the second day of the study once all the tasks for the day were completed. The participant was required to finish answering these before beginning tasks on the next day to ensure that their answers were relevant to that day only. 
\begin{enumerate}
    \item Is there something that you would improve about the interface so far?
    \item How would you compare HAT to other control interfaces?
    \item What are your general thoughts on HAT that you haven't said already? Other thoughts?
\end{enumerate}
\subsubsection{Day 3 Questions} 
These were asked at the end of the third day of the study once all the tasks for the day were completed. The participant was required to finish answering these before beginning tasks on the next day to ensure that their answers were relevant to that day only. 
\begin{enumerate}
    \item How do you like controlling the robot using HAT through line-of-sight versus through the camera feeds?
    \item How does HAT compare with a web interface for tasks that require camera views? What are the pros and cons of each control interface?  
    \item What are your thoughts on the compatibility of the HAT interface with both your robot and your computer?
    \item What are the pros and cons of using your current head tracking interface versus using the HAT interface?  
    \item Are there any challenges in transitioning between the robot and your computer? Why?  
    \item What are your general thoughts on HAT that you haven't said already? Other thoughts?
\end{enumerate}
\subsubsection{Day 4 Questions} 
These were asked at the end of the fourth day of the study once all the tasks for the day were completed. The participant was required to finish answering these before beginning tasks on the next day to ensure that their answers were relevant to that day only. 
\begin{enumerate}
    \item How would you compare your overall workload when using HAT versus other control interfaces?
    \item What are your initial thoughts on Driver Assistance?
    \item What are your general thoughts on HAT that you haven't said already? Other thoughts?
\end{enumerate}
\subsubsection{Day 5 Questions} 
These were asked at the end of the fifth day of the study once all the tasks for the day were completed. The participant was required to finish answering these before beginning tasks on the next day to ensure that their answers were relevant to that day only. 
\begin{enumerate}
    \item Are there any tasks out of the ones we have done so far that you prefer doing with the robot and HAT interface over asking your caregiver to do them for you?  
    \item Is there a difference in the independence that you feel after using the HAT interface? Why?
    \item What are your thoughts on Driver Assistance? Anything you haven't shared already?
    \item Do you still feel like you are in control when using Driver Assistance? Why or why not?
    \item What other tasks do you see Driver Assistance being useful for?
\end{enumerate}
\subsubsection{Day 6 Questions} 
These were asked at the end of the sixth day of the study once all the tasks for the day were completed. The participant was required to finish answering these before beginning tasks on the next day to ensure that their answers were relevant to that day only.
\begin{enumerate}
    \item What are tasks that you would prefer using the HAT for and tasks that you would prefer using a web interface for? Why?
    \item Did the HAT interface offer any additional features or functionalities that you found beneficial or unique compared to other assistive interfaces that you have used in the past?  
    \item Any new insights about HAT from the open day of testing (cup grabbing, wiping table, fridge opening, drawer opening)?
\end{enumerate}
\subsubsection{Day 7 and End of Study Questions} 
These were asked at the end of the last day of the study once all the tasks for the day were completed. The participant was given a week to answer these questions to ensure the evaluation of the HAT interface as a whole.
\begin{enumerate}
    \item How has your impression of HAT changed after using it for a week?
    \item Is there something that you would improve about the interface?
    \item Were there any notable barriers or limitations you encountered while using the interface?
    \item Do you enjoy the physical design of the interface? Is it comfortable to wear?
    \item Do you think you improved at using HAT over time with more practice? 
    \item Assuming no technical issues with HAT or the robot, would you use HAT to do certain tasks instead of calling your caregiver? What would they be? 
    \item Assuming no technical issues with HAT or the robot, would you feel safe using HAT?
    \item Is the HAT interface intuitive to use in general? Are there specific interface commands (head tilts/mouse clicking) that are more intuitive than others?
    \item Was the HAT interface efficient during full teleoperation? How does the HAT interface compare to a web interface in terms of efficiency?
\end{enumerate}

%\clearpage
%\pagenumbering{arabic} 

%\setcounter{section}{1} % Reset the section counter to 0
%\renewcommand{\thesection}{S\arabic{section}} % Redefine the section counter format
%\setcounter{subsection}{1} % Reset the subsection counter to 0
%\renewcommand{\thesubsection}{S\arabic{section}.\arabic{subsection}} % Redefine the subsection counter format

\subsection{Task Setup and Description}
\label{tasksetup}
\subsubsection{Predefined Tasks}
\begin{enumerate}
    \item Blanket+Tissue+Trash \\
    $\bullet$ Setup: Henry was laying in his bed for this task. He was covered waist down with a blanket. A box of tissues was placed on a table at the end of his bed and a trash can was placed against the wall to the left of his bed. \\
    $\bullet$ Description: Henry had to drive the robot towards himself and remove the blanket such that it was below his knees. Then, he had to pull a tissue from the tissue box and use it to wipe his face. Lastly, he had to throw the tissue into the trash can. The task was considered finished once the tissue was in the trash can and the HAT was back into idle mode.
    \item Docking Robot \\
    $\bullet$ Setup: Henry was sitting in his wheelchair for this task. A monitor with the GUI displayed on it was placed in front of him. The dock was placed in the hallway outside his room and the robot was placed facing the dock, about 10 feet away from it. \\
    $\bullet$ Description: Henry had to drive the robot to the dock, align the robot, and then drive the robot onto the dock. The task was considered finished once the robot was on the dock and the HAT was back into idle mode.
    \item Scratching Itch\\
    $\bullet$ Setup: Henry was laying in his bed for this task. A hairbrush placed on a stand was kept on an overbed table in front of him. The hairbrush was modified to have a larger handle to facilitate easier grasping by the robot. \\
    $\bullet$ Description: Henry had to drive the robot towards the table and pick the hairbrush up. He then had to scratch any part of his face that felt itchy a few times. The task was considered finished once he was done itching and he moved the robot arm away from his face. 
    \item Feeding\\
    $\bullet$ Setup: Henry was laying in his bed for this task. A soft spoon placed on a stand and a bowl containing a mixture of yogurt and cashew butter was kept on an overbed table in front of him. The spoon was modified to have a larger handle to facilitate easier grasping by the robot. \\
    $\bullet$ Description: Henry had to drive the robot towards the table and pick the spoon up. He then had to scoop up some food from the bowl and eat it. The task was considered finished once he put the first scoop of food in his mouth and he put the HAT back in idle.
    \item Removing Soiled Towel\\
    $\bullet$ Setup: Henry was sitting in his wheelchair the first time this task was conducted and laying in his bed the second time. A towel was placed on his bed and a laundry basket was kept against the wall to the left of his bed. A monitor with the GUI displayed on it was placed in front of him.\\
    $\bullet$ Description: Henry had to drive the robot towards the towel and grasp it. He then had to drive the robot to the laundry basket and drop the towel into it. The task was considered finished once the towel was in the laundry basket and he put the HAT back into idle.
    \item Turning Lights On\\
    $\bullet$ Setup: Henry was laying in his bed for this task. A 3D printed light switch adaptor that could be pushed upwards to turn the light switch on was attached to the switch on the wall behind Henry’s bed. A monitor with the GUI displayed on it was placed in front of him.\\
    $\bullet$ Description: Henry had to drive the robot over to the light switch and turn it on. The task was considered finished once the lights were on and he put the HAT back into idle.
    \item Applying Lotion\\
    $\bullet$ Setup: Henry was laying in his bed for this task. A plate with some lotion and a lotion application tool was placed on a table at the end of Henry’s bed. The tool consisted of a block of foam wrapped in a towel. \\
    $\bullet$ Description: Henry first had to drive the robot towards himself and remove the blanket on him such that his legs were completely uncovered. Next, he had to drive the robot towards the table, grasp the lotion application tool and dip it in the plate of lotion. Lastly, he had to drive the robot over to himself and move the lotion application tool against his leg several times. The task was considered finished once he applied lotion on his leg put the HAT back into idle. For the calculation of task time, we only considered the lotion application part of the task, not the blanket removal. 
    \item Manipulating Blinds\\
    $\bullet$ Setup: Henry was laying in his bed for this task. A monitor with the GUI displayed on it was placed in front of him. The blind cord handles were enlarged to facilitate easier grasping by the robot.  \\
    $\bullet$ Description: Henry had to drive the robot over to the blinds and pull down on them. The task was considered finished once the blinds were halfway up and he put the HAT back into idle.
\end{enumerate}

\subsubsection{Driver Assistance Tasks}
\begin{enumerate}
    \item Cleaning Up Cups \\
    $\bullet$ Setup: Henry was sitting in his wheelchair in the kitchen for this task. Two cups were placed on the dining table in front of Henry about 4 feet from each other. \\
    $\bullet$ Description: Henry had to first drive the robot to the cup placed further away from it, grasp it and place it at the end of the table. He then repeated this for the other cup. The task was considered finished once both the cups were at the end of the table and he put the HAT back into idle.
    \item Fetching Red Bull \\
    $\bullet$ Setup: Henry was sitting in his wheelchair in the kitchen for this task. A can of Red Bull was placed in a cluttered environment on one end of the dining table. \\
    $\bullet$ Description: Henry had to drive the robot over to the can of Red Bull, grasp it and then place it at the other end of the dining table. The task was considered finished once he put the HAT back into idle.
    \item Removing Soiled Towel\\
    $\bullet$ Setup: Henry was laying in his bed for this task. A towel was placed on Henry’s bed. A laundry basket was placed against the wall to the left of Henry’s bed. \\
    $\bullet$ Description: Henry had to drive the robot towards the towel on the bed, grasp it and put it in the laundry basket. The task was considered finished once the towel was in the laundry basket and he put the HAT back into idle.
\end{enumerate}

\subsubsection{Open Tasks}
\begin{enumerate}
    \item Playing Cards\\
    $\bullet$ Setup: Henry was laying in his bed for this task. A card wheel that could be spun by linearly moving a handle was placed on an overbed table in front of Henry. The robot was placed next to the bed. \\
    $\bullet$ Description: Henry and his wife played a couple of games of poker. Henry would tell his wife the number of cards he wished to discard by blinking an appropriate number of times. He would then spin the card wheel such that the card at the top position was the one he wanted to discard.
    \item Opening Drawer+Fetching Tape\\
    $\bullet$ Setup: Henry was in his wheelchair in the kitchen for the task. A monitor with the GUI displayed on it was placed in front of him. A foam block was taped to the handle of the drawer in the living room adjacent to the kitchen to make it possible to grasp. The robot was placed in the kitchen next to Henry. \\
    $\bullet$ Description: Henry had to drive the robot over to the living room, open the drawer and grasp a roll of tape that was already inside it. He then had to drive the robot back to the kitchen and place the roll of tape in front of him. The task was considered finished once the roll of tape was in front of him and he put the HAT back into idle.
    \item Grasping Towel+Wiping Table\\
    $\bullet$ Setup: Henry was in his wheelchair in the kitchen for the task. A towel was placed on a shelf a few feet away from him. There were a few drops of water on the table in front of him. Henry was allowed to use Driver Assistance. \\
    $\bullet$ Description: Henry had to drive the robot over to the shelf and grasp the towel using Driver Assistance. He then had to drive the robot back to the table and wipe the water on it. The task was considered finished once no water was on the table and he put the HAT back into idle.
    \item Grasping+Fetching Bottle\\
    $\bullet$ Setup: Henry was in his wheelchair in the kitchen for the task. A toddler training cup was placed on a shelf a few feet away from him. Henry was allowed to use Driver Assistance. \\
    $\bullet$ Description: Henry had to drive the robot over to the shelf and grasp the cup using Driver Assistance. He then had to drive the robot back towards himself and place it on the armrest of his chair. The task was considered finished once the cup was on the armrest and he put the HAT back into idle.
    \item Opening Fridge\\
    $\bullet$ Setup: Henry was in his wheelchair in the kitchen for the task. A monitor with the GUI displayed on it was placed in front of him. A foam block was taped to the handle of the mini fridge on the other end of the kitchen to make it possible to grasp. \\
    $\bullet$ Description: Henry had to drive the robot over to the mini fridge and open it. The task was considered finished once the fridge door was open and he put the HAT back into idle.
\end{enumerate}

   \begin{figure*}[t!]
      \centering
      \includegraphics[width = \textwidth]{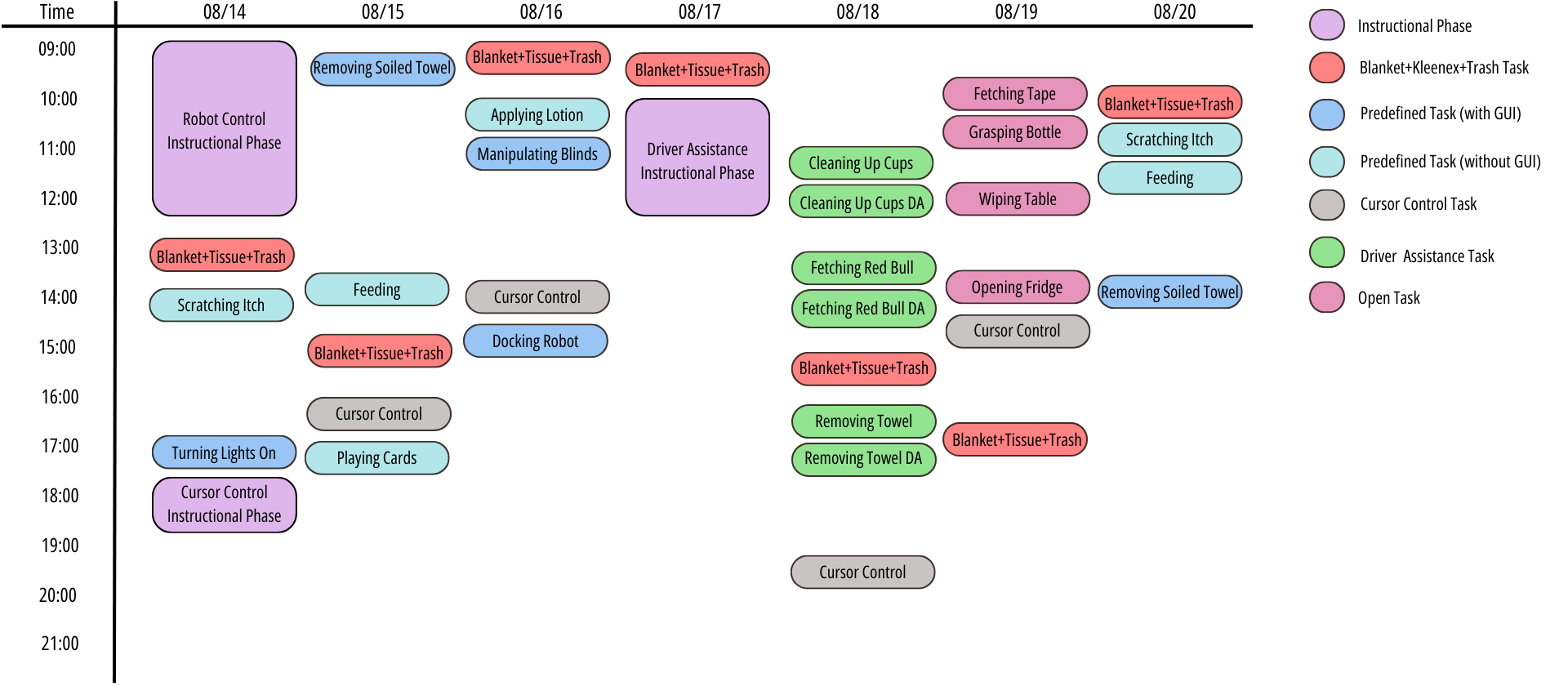}
      \vspace{-0.4cm}
      \caption{A comprehensive human study calendar showcasing instructional sessions and all tasks conducted. (DA stands for Driver Assistance.)}
      \vspace{-0.4cm}
      \label{calendar}
   \end{figure*}

\newcolumntype{C}[1]{>{\arraybackslash}p{#1}}
\newcolumntype{Y}{>{\centering\arraybackslash}X}

\begin{table*}
  \centering
  \caption{Predefined Tasks}
  \label{tab:sup1}
\begin{tabularx}{\textwidth}{ |l| *{1}{Y} |c| *{10}{Y|} }

\toprule
            Task Name & Day & Time* & Resets & L1 & L2 & L3 &  L4 &  L5 & L6 &  L7 \\
\midrule
 Blanket+Tissue+Trash &   1 & 16:30 &      1 &  6 &  3 &  6 & - & - &  4 & - \\
      Scratching Itch &   1 & 08:49 &      0 &  6 &  3 &  6 & - & - &  4 & - \\
    Turning Lights On &   1 & 07:14 &      0 &  5 &  6 &  6 & - & - &  4 &   3 \\
Removing Soiled Towel &   2 & 08:01 &      0 &  3 &  3 &  5 & - & - &  2 &   4 \\
              Feeding &   2 & 09:08 &      1 &  6 &  5 &  6 &   5 &   4 &  4 & - \\
 Blanket+Tissue+Trash &   2 & 14:43 &      1 &  6 &  3 &  6 &   5 &   4 &  4 & - \\
 Blanket+Tissue+Trash &   3 & 10:42 &      0 &  6 &  6 &  7 &   5 &   5 &  5 & - \\
   Applying Lotion &   3 & 07:40 &      0 &  6 &  6 &  6 &   5 &   5 &  3 & - \\
  Manipulating Blinds &   3 & 08:45 &      1 &  7 &  6 &  6 &   6 &   6 &  5 &   1 \\
        Docking Robot &   3 & 02:31 &      0 &  6 &  5 &  6 &   5 &   5 &  4 &   1 \\
 Blanket+Tissue+Trash &   4 & 09:17 &      0 &  6 &  5 &  6 &   5 &   5 &  4 & - \\
 Blanket+Tissue+Trash &   5 & 12:55 &      0 &  5 &  5 &  6 &   4 &   5 &  4 & - \\
 Blanket+Tissue+Trash &   6 & 13:51 &      0 &  6 &  6 &  6 &   5 &   5 &  5 & - \\
 Blanket+Tissue+Trash &   7 & 08:44 &      0 &  7 &  6 &  6 &   5 &   5 &  4 & - \\
      Scratching Itch &   7 & 05:17 &      0 &  6 &  6 &  6 &   5 &   5 &  4 & - \\
              Feeding &   7 & 07:18 &      0 &  6 &  6 &  6 &   5 &   5 &  5 & - \\
Removing Soiled Towel &   7 & 04:26 &      0 &  6 &  6 &  6 &   5 &   5 &  5 &   1 \\
\midrule
               median & - &   - &    - &  6 &  6 &  6 &   5 &   5 &  4 &   1 \\
                  IQR & - &   - &      - &  0 &  1 &  0 &   0 &   0 &  1 &   2 \\
\bottomrule

   \multicolumn{11}{C{17cm}}{L1 (7 point Likert Item): The control interface was easy to use for the previous task.} \\ 
   \multicolumn{11}{C{17cm}}{L2 (7 point Likert Item): I was able to complete the previous task without any errors using the control interface.} \\
   \multicolumn{11}{C{17cm}}{L3 (7 point Likert Item): If there were errors while performing the previous task, the control interface allowed easy recovery.} \\
   \multicolumn{11}{C{17cm}}{L4(7 point Likert Item): HAT enabled control of the robot better than the web interface for the previous task.} \\
   \multicolumn{11}{C{17cm}}{L5 (7 point Likert Item): I would prefer to use HAT over the web interface for the previous task.} \\
   \multicolumn{11}{C{17cm}}{L6 (7 point Likert Item): The control interface enabled control of the robot in a reasonable amount of time for the previous task.} \\
   \multicolumn{11}{C{17cm}}{L7 (7 point Likert Item):The control interface was harder to use without line of sight than with line of sight.}  \\
   \multicolumn{11}{C{17cm}}{*Task times are listed in minutes and seconds.}
\end{tabularx}
\end{table*}

\begin{table*}
  \centering
  \caption{Open Tasks}
  \label{tab:sup2}
\begin{tabularx}{\textwidth}{ |l| *{1}{Y} |c| *{16}{Y|} }
\toprule
                     Task Name & Day & Time* & R** & DA** & L1 & L2 & L3 &  L4 &  L5 & L6 &  L7 &  L8 &  L9 & L10 & L11 & L12 \\
\midrule
                     Playing Cards &   2 &   - &    - &        - &  7 &  6 &  7 &   6 &   6 &  6 & - & - & - & - & - & - \\
Opening Drawer+Fetching Tape &   6 &   - &      2 &        - &  3 &  5 &  6 &   5 &   5 &  1 &   1 & - & - & - & - & - \\
   Grasping+Fetching Bottle &   6 & 04:28 &      0 & \checkmark &  6 &  6 &  6 & - & - &  5 & - &   7 &   7 &   7 &   7 &   7 \\
Grasping Towel+Wiping Table &   6 & 05:46 &      0 & \checkmark &  6 &  4 &  6 & - & - &  6 & - &   7 &   7 &   7 &   7 &   7 \\
                Opening Fridge &   6 & 09:18 &      0 &        - &  6 &  6 &  6 &   5 &   5 &  3 &   1 & - & - & - & - & - \\
\bottomrule
    \multicolumn{17}{C{17cm}}{L1 (7 point Likert Item): The control interface was easy to use for the previous task.} \\ 
   \multicolumn{17}{C{17cm}}{L2 (7 point Likert Item): I was able to complete the previous task without any errors using the control interface.} \\
   \multicolumn{17}{C{17cm}}{L3 (7 point Likert Item): If there were errors while performing the previous task, the control interface allowed easy recovery.} \\
   \multicolumn{17}{C{17cm}}{L4 (7 point Likert Item): HAT enabled control of the robot better than the web interface for the previous task.} \\
   \multicolumn{17}{C{17cm}}{L5 (7 point Likert Item): I would prefer to use HAT over the web interface for the previous task.} \\
   \multicolumn{17}{C{17cm}}{L6 (7 point Likert Item): The control interface enabled control of the robot in a reasonable amount of time for the previous task.} \\
   \multicolumn{17}{C{17cm}}{L7 (7 point Likert Item):The control interface was harder to use without line of sight than with line of sight.} \\
   \multicolumn{17}{C{17cm}}{L8 (7 point Likert Item): I felt like I was in control while using driver assistance for the previous task.}  \\
   \multicolumn{17}{C{17cm}}{L9 (7 point Likert Item): The robot continued to do what I wanted it to in driver assistance mode for the previous task.}  \\
   \multicolumn{17}{C{17cm}}{L10 (7 point Likert Item): I was able to accomplish the previous task more efficiently using driver assistance mode. }  \\
   \multicolumn{17}{C{17cm}}{L11 (7 point Likert Item): Driver assistance reduced frustration that I would feel when doing the same task using full teleoperation for the previous task.}  \\
   \multicolumn{17}{C{17cm}}{L12 (7 point Likert Item): I prefer using driver assistance over full teleoperation for the previous task.}  \\
   \multicolumn{17}{C{17cm}}{*Task times are listed in minutes and seconds.} \\
   \multicolumn{17}{C{17cm}}{** R stands for Resets and DA stands for Driver Assistance} \\
   \multicolumn{17}{C{17cm}}{***The prismatic robot arm lacked the necessary torque to completely open the drawer. Thus, we reduced the task time by 13:29, which is the time Henry spent repeatedly attempting to grasp the tape before a researcher intervened to facilitate access to the tape.}
\end{tabularx}
\end{table*}

\begin{table*}
  \centering
  \caption{Driver Assistance Tasks}
  \label{tab:sup3}
\begin{tabularx}{\textwidth}{ |c| *{15}{Y|} }
\toprule
                  Task Name & Day & Time* & R** &       DA** & L1 & L2 & L3 &  L4 &  L5 & L6 &  L7 &  L8 &  L9 & L10 \\
\midrule
                Cleaning Cups &   5 & 07:34 &      1 &        - &  6 &  5 &  6 &   5 &   5 &  4 & - & - & - & - \\
          Cleaning Cups(DA**) &   5 & 07:33 &      0 & \checkmark &  7 &  6 &  7 & - & - &  6 &   7 &   6 &   7 &   7 \\
    Fetching Red Bull (DA**) &   5 & 03:18 &      0 & \checkmark &  7 &  7 &  7 & - & - &  7 &   7 &   6 &   7 &   7 \\
     Fetching    Red Bull  &   5 & 11:03 &      0 &        - &  2 &  3 &  3 &   4 &   4 &  2 & - & - & - & - \\
      Removing Soiled Towel &   5 & 08:47 &      0 &        - &  5 &  5 &  6 &   4 &   5 &  4 & - & - & - & - \\
Removing Soiled Towel(DA**) &   5 & 04:52 &      0 & \checkmark &  6 &  6 &  6 & - & - &  5 &   6 &   7 &   7 &   7 \\
\bottomrule
    \multicolumn{15}{C{17cm}}{L1 (7 point Likert Item): The control interface was easy to use for the previous task.} \\ 
   \multicolumn{15}{C{17cm}}{L2 (7 point Likert Item): I was able to complete the previous task without any errors using the control interface.} \\
   \multicolumn{15}{C{17cm}}{L3 (7 point Likert Item): If there were errors while performing the previous task, the control interface allowed easy recovery.} \\
   \multicolumn{15}{C{17cm}}{L4 (7 point Likert Item): HAT enabled control of the robot better than the web interface for the previous task.} \\
   \multicolumn{15}{C{17cm}}{L5 (7 point Likert Item): I would prefer to use HAT over the web interface for the previous task.} \\
   \multicolumn{15}{C{17cm}}{L6 (7 point Likert Item): The control interface enabled control of the robot in a reasonable amount of time for the previous task.} \\
   \multicolumn{15}{C{17cm}}{L7 (7 point Likert Item): I felt like I was in control while using driver assistance for the previous task.}  \\
   \multicolumn{15}{C{17cm}}{L8 (7 point Likert Item): The robot continued to do what I wanted it to in driver assistance mode for the previous task.}  \\
   \multicolumn{15}{C{17cm}}{L9 (7 point Likert Item): I was able to accomplish the previous task more efficiently using driver assistance mode. }  \\
   \multicolumn{15}{C{17cm}}{L10 (7 point Likert Item): Driver assistance reduced frustration that I would feel when doing the same task using full teleoperation for the previous task.}  \\
   \multicolumn{15}{C{17cm}}{L11 (7 point Likert Item): I prefer using driver assistance over full teleoperation for the previous task.}  \\
   \multicolumn{15}{C{17cm}}{*Task times are listed in minutes and seconds.} \\
   \multicolumn{15}{C{17cm}}{**R stands for Resets and DA stands for Driver Assistance}
\end{tabularx}

\begin{tabularx}{\textwidth}{ |c| *{6}{Y|} }
\toprule
                  Task Name & TLX1 & TLX2 & TLX3 & TLX4 & TLX5 & TLX6 \\
\midrule
                Cleaning Up Cups &    3 &    1 &    1 &    3 &    3 &    2 \\
          Cleaning Up Cups(DA**) &    1 &    1 &    1 &    2 &    1 &    1 \\
    Fetching Red Bull (DA**) &    1 &    1 &    1 &    1 &    1 &    1 \\
      Fetching Red Bull  &    5 &    2 &    2 &    3 &    5 &    3 \\
      Removing Soiled Towel &    5 &    2 &    2 &    3 &    5 &    3 \\
Removing Soiled Towel(DA**) &    1 &    1 &    1 &    2 &    1 &    1 \\
\bottomrule
\multicolumn{7}{C{17cm}}{TLX1 (7 point NASA TLX Scale): How mentally demanding was the task?} \\
\multicolumn{7}{C{17cm}}{TLX2 (7 point NASA TLX Scale): How physically demanding was the task?} \\
\multicolumn{7}{C{17cm}}{TLX3 (7 point NASA TLX Scale): How rushed or hurried was the pace of the task?} \\
\multicolumn{7}{C{17cm}}{TLX4 (7 point NASA TLX Scale): How successful do you think you were in accomplishing what you were asked to do?} \\ \multicolumn{7}{C{17cm}}{TLX5 (7 point NASA TLX Scale): How hard did you have to work to accomplish your level of performance?} \\
\multicolumn{7}{C{17cm}}{TLX6 (7 point NASA TLX Scale): How insecure, discouraged, irritated, stressed, and annoyed were you?} \\
\multicolumn{7}{C{17cm}}{**DA stands for Driver Assistance}

\end{tabularx}

\end{table*}

\begin{table*}
  \centering
  \caption{Daily Responses}
  \label{tab:sup4}
\begin{tabularx}{\textwidth}{ |c| *{19}{Y|} }
\toprule
   Day & L1 &  L2 &  L3 & L4 & L5 &  L6 &  L7 &  L8 &  L9 & L10 & L11 & L12 & TLX1 & TLX2 & TLX3 & TLX4 & TLX5 & TLX6 \\
\midrule
     1 &  6 &   4 &   5 &  6 &  3 & - & - & - & - & - & - & - &    6 &    2 &    2 &    2 &    5 &    2 \\
     2 &  5 &   2 &   5 &  6 &  2 &   4 &   4 & - & - & - & - & - &    6 &    1 &    2 &    3 &    5 &    1 \\
     3 &  5 &   4 &   5 &  6 &  2 &   5 &   5 & - & - & - & - & - &    3 &    2 &    1 &    2 &    3 &    1 \\
     4 &  6 &   4 &   6 &  6 &  2 &   5 &   5 & - & - & - & - & - &    3 &    2 &    1 &    1 &    3 &    1 \\
     5 &  5 &   5 &   5 &  6 &  2 &   5 &   5 &   7 &   7 &   7 &   7 &   7 &    3 &    1 &    2 &    2 &    3 &    2 \\
     6 &  6 &   4 &   5 &  6 &  1 &   6 &   6 &   7 &   7 &   7 &   7 &   7 &    6 &    2 &    2 &    3 &    3 &    2 \\
     7 &  6 &   6 &   6 &  6 &  2 &   5 &   5 & - & - & - & - & - &    2 &    1 &    1 &    1 &    2 &    1 \\
\midrule
median &  6 &   4 &   5 &  6 &  2 &   5 &   5 &   7 &   7 &   7 &   7 &   7 &    3 &    2 &    2 &    2 &    3 &    1 \\
   IQR &  1 & 0.5 & 0.5 &  0 &  0 &   0 &   0 &   0 &   0 &   0 &   0 &   0 &    3 &    1 &    1 &    1 &    1 &    1 \\
\bottomrule
    \multicolumn{19}{C{17cm}}{L1 (7 point Likert Item): The control interface was easy to use today.} \\ 
    \multicolumn{19}{C{17cm}}{L2 (7 point Likert Item): The control interface enabled control of the robot in a reasonable amount of time today.} \\
   \multicolumn{19}{C{17cm}}{L3 (7 point Likert Item): I was able to complete the tasks today without any errors using the control interface.} \\
   \multicolumn{19}{C{17cm}}{L4 (7 point Likert Item): If there were errors while performing the tasks today, the control interface allowed easy recovery.} \\
   \multicolumn{19}{C{17cm}}{L5 (7 point Likert Item): The control interface was harder to use without line of sight than with line of sight today.} \\
   \multicolumn{19}{C{17cm}}{L6 (7 point Likert Item): HAT enabled control of the robot better than the web interface.} \\
   \multicolumn{19}{C{17cm}}{L7 (7 point Likert Item): I would prefer to use HAT over the web interface.} \\
   \multicolumn{19}{C{17cm}}{L8 (7 point Likert Item): I felt like I was in control while using driver assistance today.}  \\
   \multicolumn{19}{C{17cm}}{L9 (7 point Likert Item): The robot continued to do what I wanted it to in driver assistance mode today.}  \\
   \multicolumn{19}{C{17cm}}{L10 (7 point Likert Item): I was able to accomplish tasks today more efficiently using driver assistance mode. }  \\
   \multicolumn{19}{C{17cm}}{L11 (7 point Likert Item): Driver assistance reduced frustration that I would feel when doing the same tasks using full teleoperation.}  \\
   \multicolumn{19}{C{17cm}}{L12 (7 point Likert Item): I prefer using driver assistance over full teleoperation for the tasks today.}  \\
    \multicolumn{19}{C{17cm}}{TLX1 (7 point NASA TLX Scale): How mentally demanding was the task?} \\
    \multicolumn{19}{C{17cm}}{TLX2 (7 point NASA TLX Scale): How physically demanding was the task?} \\
    \multicolumn{19}{C{17cm}}{TLX3 (7 point NASA TLX Scale): How rushed or hurried was the pace of the task?} \\
    \multicolumn{19}{C{17cm}}{TLX4 (7 point NASA TLX Scale): How successful do you think you were in accomplishing what you were asked to do?} \\ \multicolumn{19}{C{17cm}}{TLX5 (7 point NASA TLX Scale): How hard did you have to work to accomplish your level of performance?} \\
    \multicolumn{19}{C{17cm}}{TLX6 (7 point NASA TLX Scale): How insecure, discouraged, irritated, stressed, and annoyed were you?} \\
\end{tabularx}

\end{table*}

\clearpage

\end{document}